\documentclass[11pt,a4paper]{article}

\usepackage[margin=1in]{geometry}

\usepackage[numbers,sort&compress]{natbib}
\usepackage{booktabs}
\usepackage{multirow}
\usepackage{array}
\usepackage{graphicx}
\usepackage{amssymb}
\usepackage{amsmath}
\usepackage{textcomp}
\usepackage[section]{placeins}
\usepackage{float}
\usepackage{url}
\usepackage[capitalize]{cleveref}
\graphicspath{{figures/}}

\begin{document}

\title{AnchorScore: A CLIP-Based Diagnostic of MLLM Annotation Difficulty}

\author{Yan Ma\thanks{E-mail: mayan@csust.edu.cn} \\
  \small School of Foreign Studies, Changsha University of Science and Technology, Changsha, China\\
  \small and School of Education, Hunan Agricultural University, Changsha, China
\and
Lizhuo Zhang\thanks{Corresponding author. E-mail: zhanglizhuo@hunau.edu.cn} \\
  \small School of Information and Intelligence, Hunan Agricultural University, Changsha, China}

\begin{abstract}
Multimodal large language models (MLLMs) are widely used for automated annotation, yet their per-class accuracy varies widely (e.g., 12\%--98\% across the 13 classes of three classroom sub-datasets, 8$+$3$+$2) and is expensive to measure: evaluating one 27B MLLM on 5{,}416 validation images takes roughly 14 hours, whereas a frozen-CLIP pass over the same images completes in about 3 minutes. A low-cost signal for ranking classes by expected MLLM annotation difficulty \emph{a priori} remains underexplored. Building on the AnchorProxy construct (per-class zero-shot CLIP accuracy) introduced in the companion study, this paper systematically evaluates its full-frame formulation---termed \textbf{AnchorScore} here---as an \emph{a priori} diagnostic that flags the classes MLLMs are least likely to annotate reliably.

On classroom behavior data (SCB5, 13 classes, 6 MLLMs), AnchorScore correlates with per-class MLLM accuracy (Spearman $\rho=0.769$, $p=0.002$, $n=13$, with a wide 95\% CI reflecting the small class count). None of the alternative difficulty predictors (DINOv2, ResNet-50, SigLIP, or MLLM self-verbalized uncertainty) showed a significant class-level correlation at $n=13$, though several operate near their accuracy floors. A cross-model consensus control suggests AnchorScore primarily captures a shared class-difficulty factor rather than a CLIP-specific signal: conditioning on the consensus of the other five MLLMs leaves a pooled partial correlation near zero ($\hat{\rho}=0.067$), with a wide 95\% CI that cannot exclude a substantial CLIP-specific component. A prompt-randomization control (replicated on both SCB5 and Stanford40) further shows the association depends on class--prompt alignment---it collapses when class descriptions are shuffled or replaced---arguing against a pure image-side difficulty account. An independent replication on Stanford40 Actions yields a nearly identical effect ($\rho=0.817$, $p<0.001$, 95\% CI $[0.622, 0.930]$ at $n=40$); the association is strongest on activity-recognition data and attenuates on medical and satellite imagery.

Three practical applications follow: a deployable hybrid CLIP/MLLM routing strategy (predicted-class routing: up to $+$23~pp over CLIP-only at roughly 44\% MLLM cost savings under the benchmark's canonical MLLM evaluation; realized runs deliver $+$21.7~pp under that protocol and $+$17.5~pp under the full-frame deployment condition), prompt disambiguation on hard classes (exploratory), and review-priority prediction for human verification. AnchorScore does not estimate exact MLLM accuracy; it provides a low-cost ranking signal that directs expensive MLLM evaluation to the classes where it is most informative.
\end{abstract}

\maketitle


\section{Introduction}
\label{sec:intro}

Multimodal large language models (MLLMs) are now widely adopted for automatic data annotation~\cite{liu2023visual}. Their per-class accuracy, however, varies dramatically. On a classroom behavior dataset (SCB5)~\cite{ma2026agreement}, annotation accuracy averaged across six MLLMs in the evaluation exceeds 97\% for ``blackboard-writing'' but reaches only 12\% for ``answer'', a gap of 85~pp (percentage points). Without running the full annotation pipeline on tens of billions of parameters, how can practitioners identify which classes an MLLM will annotate reliably?

\begin{figure}[t]
\centering
\includegraphics[width=\columnwidth]{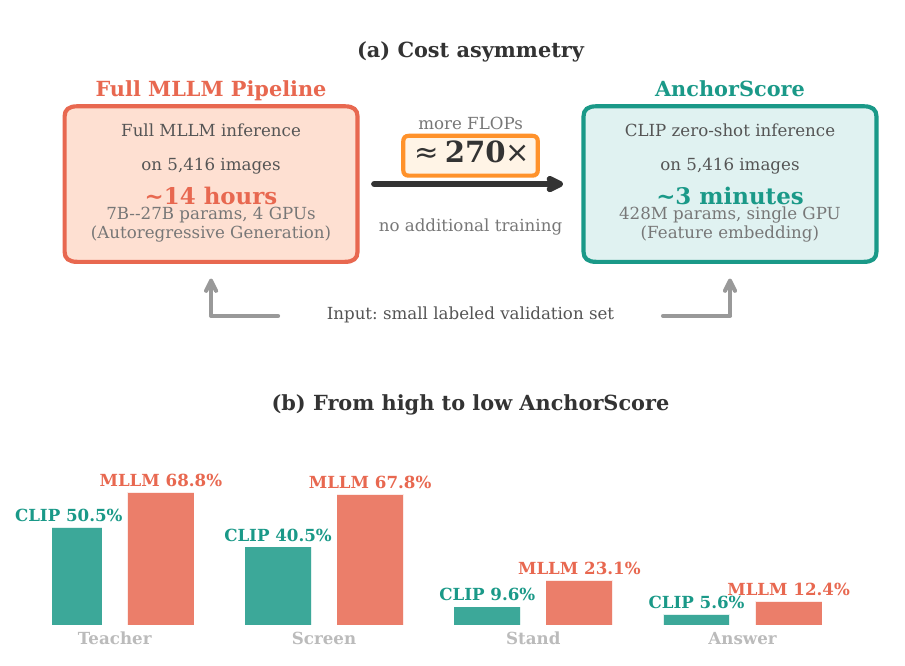}
\caption{AnchorScore at a glance. (a)~Cost asymmetry: evaluating one 27B MLLM on the 5{,}416 SCB5 validation images takes roughly 14 hours, whereas a single frozen-CLIP forward pass yields per-class scores in about 3 minutes ($\approx$270$\times$ fewer FLOPs, no additional training). (b)~Illustrative classes ranked by AnchorScore: per-class CLIP zero-shot accuracy (left bars) and six-MLLM mean accuracy (right bars) co-move from ``teacher'' (high AnchorScore) to ``answer'' (very low AnchorScore); values are the canonical per-class estimates of \S\ref{sec:main_result}.}
\label{fig:concept}
\end{figure}

Existing approaches, such as running a pilot sample through the MLLM or extracting model-specific uncertainty~\cite{kadavath2022language}, remain computationally expensive. Evaluating one 27B MLLM on 5{,}416 validation images takes roughly 14 hours, while requiring access to the model itself. This paper evaluates a lightweight alternative: \textbf{AnchorScore}, defined as the per-class zero-shot accuracy of a frozen CLIP model~\cite{radford2021clip}. If CLIP's zero-shot accuracy on a class is low, will a larger MLLM (26--35B parameters) also struggle? Given a small labeled validation set (as few as 20 images per class), a single CLIP forward pass (about 3 minutes for the 5{,}416-image validation set) can reveal which classes are likely to challenge the MLLM, producing a per-class difficulty map that would otherwise require hours of MLLM inference. \Cref{fig:concept} illustrates the cost asymmetry and the anchor intuition.

Validation on SCB5 (13 classes, 6 MLLMs) shows a significant class-level correlation between AnchorScore and MLLM accuracy ($\rho=0.769$, $p=0.002$). Alternative probes (SigLIP, DINOv2, ResNet-50, MLLM self-uncertainty) show no significant class-level correlation. Nor do the two most obvious cheap alternatives: a 7B MLLM pilot on the same labeled images carries no class-difficulty signal ($\rho=-0.03$; \S\ref{sec:extended_baselines}), and routing by CLIP's own prediction confidence performs no better than random (\S\ref{sec:hybrid})---the cross-architecture CLIP signal is what remains. A consensus control indicates the signal is consistent with a shared class-difficulty interpretation rather than a CLIP-specific mechanism. An independent replication on Stanford40 Actions (40 classes, $\rho=0.817$) corroborates the effect. Three practical applications follow from this diagnostic, together forming a resource-aware annotation workflow: hybrid CLIP/MLLM annotation routing, prompt disambiguation guided by CLIP confusion patterns, and review-priority ranking.

AnchorScore is not a calibrated accuracy estimator; it is a low-cost diagnostic for ranking classes by relative difficulty. The contributions are threefold:
\begin{itemize}
\item \textbf{Diagnostic discovery.} This study shows that a cheap, training-free probe from a \emph{different} model family---frozen-CLIP per-class zero-shot accuracy---predicts where MLLM annotation is unreliable, correlating with per-class MLLM accuracy on SCB5 ($\rho=0.769$, $p=0.002$) and replicating on Stanford40 ($\rho=0.817$, $p<0.001$). The value of the probe is precisely that the signal is obtainable before any MLLM inference, turning MLLM evaluation from a per-model cost into an allocatable budget.
\item \textbf{Shared-difficulty characterization.} AnchorScore is consistent with a \emph{shared} class-difficulty interpretation rather than a CLIP-specific mechanism: alternative predictors (SigLIP, DINOv2, ResNet-50, MLLM self-uncertainty) show no significant class-level correlation, a cross-model consensus control leaves no significant residual association beyond the other MLLMs' agreement, and a prompt-randomization control shows the association depends on class--prompt alignment---it collapses when the alignment is broken, in both SCB5 and Stanford40---arguing against a pure image-side difficulty account.
\item \textbf{Practical uses.} The signal supports three complementary workflows: hybrid CLIP/MLLM annotation, confusion-guided prompt disambiguation, and review-priority prediction.
\end{itemize}
The contribution is empirical rather than algorithmic: AnchorScore is simply CLIP zero-shot accuracy, and its value lies in the evidence chain establishing when and where this diagnostic is reliable.

\section{Related Work}
\label{sec:related}

\textbf{CLIP for zero-shot evaluation} established that contrastive vision-language pretraining enables effective zero-shot transfer~\cite{radford2021clip}. CLIPScore~\cite{hessel2021clipscore} uses CLIP to evaluate text generation quality via image-text alignment, showing that CLIP's embedding space can serve as a reference signal---a philosophy extended here to the MLLM annotation setting. \textbf{The present work differs} by using CLIP accuracy as a diagnostic probe for a \emph{different} model family (MLLMs), not for evaluating CLIP itself.

\textbf{MLLMs as annotators} have been adopted for medical imaging, content moderation, and general-purpose labeling~\cite{liu2023visual}. On classroom behavior data~\cite{ma2026agreement}, per-class MLLM annotation reveals a bimodal accuracy distribution where visually anchored categories (e.g., \emph{blackboard-writing}) are annotated reliably while intent-dependent categories (e.g., \emph{answer}) fail universally. However, existing work does not provide \emph{a priori} per-class reliability estimates. \textbf{This work fills} this gap by showing that CLIP zero-shot accuracy identifies which classes an MLLM annotates reliably---a signal not captured by MLLM self-uncertainty or non-CLIP vision models. The closest prior, the companion study that introduced the AnchorProxy construct under a bbox-cropped protocol~\cite{ma2026agreement}, used it only as a post-hoc screening signal; the present work extends it to the full-frame protocol and supplies a systematic empirical evaluation as an \emph{a priori} predictor.

\textbf{MLLM calibration and failure detection.} Two post-hoc lines address MLLM reliability. \emph{Calibration} estimates how trustworthy the model's confidence is: logit-based methods (softmax entropy/\allowbreak max-logit)~\cite{kadavath2022language}, self-verbalized confidence~\cite{tian2023just}, and consistency-based uncertainty~\cite{kuhn2023semantic}, all of which require MLLM logit access or multiple decoding passes (verbalized confidence requires at least a full MLLM inference pass). \emph{Failure detection} instead identifies factual errors after they occur, benchmarked by CHAIR~\cite{rohrbach2018chair} and POPE~\cite{li2023pope}. \textbf{In contrast to both lines}, AnchorScore identifies difficult classes \emph{a priori} using only CLIP, without running any MLLM inference.

\textbf{Active learning} traditionally selects informative samples via uncertainty~\cite{settles2009active} or diversity~\cite{sener2018active} sampling; these signals are instance-level and are consumed \emph{inside} a training loop, where each selection depends on the current model state. \textbf{AnchorScore differs in granularity and cost}: its per-class zero-shot accuracy is a \emph{class-level} difficulty signal computed once from a frozen encoder, without iterative training or uncertainty estimation. The two address complementary questions---which \emph{images} to label next versus which \emph{classes} a downstream annotator will find difficult---and class-level difficulty can inform the per-class budget allocation that instance-level AL does not address.

\textbf{Selective prediction} allows a model to abstain from low-confidence predictions~\cite{geifman2017selective, elyaniv2010foundations}, with recent extensions to vision-language reasoning~\cite{srinivasan2024selective} and audits of its evaluation pitfalls~\cite{luth2024overcoming}. The hybrid CLIP/MLLM routing of \S\ref{sec:hybrid} belongs to this family, but differs in its signal: the abstention decision uses a \emph{class-level}, \emph{a priori} difficulty estimate from a frozen cross-modal proxy rather than instance-level predictive confidence, so it is a routing signal rather than an instance-level selective-prediction signal.

\textbf{Proxy-model approaches} such as weak-to-strong generalization (W2S)~\cite{burns2023weak} and LLM-as-judge frameworks~\cite{dubois2024lengthcontrolled, zheng2023judging} use a smaller or weaker model to assess or supervise a larger one. W2S established that a weak LM's supervision can elicit a strong LM's capabilities on NLP tasks~\cite{burns2023weak}. \textbf{The present work differs in three ways}: (a)~the proxy (CLIP, 428M) operates in a \emph{different modality space} (vision-language bridging), not as a smaller instance of the same model family, making it a genuinely cross-modal diagnostic; (b)~per-class associations are characterized, not aggregate or instance-level performance, enabling class-specific analysis for targeted interventions; (c)~the proxy requires no training, only $\geq$20 labeled images per class and 3 minutes of inference, making it immediately deployable without any MLLM inference.

\textbf{Choice of CLIP as probe.} CLIP was chosen over alternative vision-language models (e.g., SigLIP) for three reasons: its dual-encoder architecture provides direct text-image similarity, its LAION-2B training distribution plausibly overlaps with MLLM pretraining, and it is orders of magnitude lighter (428M parameters, $\sim$3 min for 5{,}416 images) than generative VLMs. This choice is validated empirically: evaluating SigLIP---a dual-encoder model trained with sigmoid pairwise loss---yields a non-significant correlation ($\rho=0.201$, $p=0.51$, $n=13$), read, with the confound caveats detailed in \S\ref{sec:extended_baselines}, as a difference in proxy quality across model families rather than evidence about CLIP's objective.

\textbf{Model-to-model performance prediction.} A related line predicts the behavior of large models from cheaper signals: scaling laws forecast aggregate performance from model and data scale~\cite{kaplan2020scaling,hoffmann2022training}, and observational scaling laws~\cite{ruan2024observational} predict emergent and agentic capabilities of large models from published benchmarks alone. AnchorScore differs in three respects: it is \emph{cross-modal} (a vision-language encoder predicting generative VLMs, not a smaller instance of the same family), it operates at the \emph{per-class} level rather than the aggregate or instance level, and it requires no training or model access---only a small labeled validation set and a single frozen-encoder pass.

\section{Method}
\label{sec:method}

The AnchorScore pipeline comprises three components---the score definition, prompt design, and implementation details---described in turn below.

\subsection{AnchorScore Definition}

AnchorScore measures the per-class zero-shot accuracy of a CLIP model on a target dataset. For a dataset with $C$ classes $\{c_1, c_2, \dots, c_C\}$ and a validation set of $N$ images, AnchorScore is computed as:

\begin{equation}
\text{AnchorScore}(c_i) = \frac{1}{N_i} \sum_{j : y_j = i} \mathbb{1}[\hat{y}_j = y_j]
\label{eq:anchorscore}
\end{equation}

where $j$ indexes the $N$ validation images, $N_i$ is the number of validation images of class $c_i$ (the label $y_j$ equals $i$ for exactly $N_i$ of the $N$ images), $y_j$ and $\hat{y}_j$ are the ground-truth label and the CLIP prediction for image $j$, both encoded as class indices in $\{1,\dots,C\}$, and $\mathbb{1}[\cdot]$ is the indicator function: it equals 1 when the CLIP prediction matches the ground-truth label and 0 otherwise. The sum therefore counts the images of class $c_i$ that CLIP classifies correctly, and $\text{AnchorScore}(c_i)$ is the fraction of such images---CLIP's per-class zero-shot accuracy. It serves as a low-cost proxy of the class-level difficulty and is the difficulty signal used throughout the correlation analyses in \Cref{sec:experiments}. The CLIP prediction is defined as:

\begin{equation}
\hat{y}_j = \arg\max_{k \in \{1,\dots,C\}} \frac{\mathbf{f}_{\text{img}}(x_j)^\top \bar{\mathbf{f}}_{\text{text}}(c_k)}{\|\mathbf{f}_{\text{img}}(x_j)\| \|\bar{\mathbf{f}}_{\text{text}}(c_k)\|}
\label{eq:clip_pred}
\end{equation}

Here $\mathbf{f}_{\text{img}}$ and $\mathbf{f}_{\text{text}}$ are the CLIP vision and text encoders and $x_j$ is the input image. Following template averaging, $\bar{\mathbf{f}}_{\text{text}}(c_k) = \frac{1}{T}\sum_{t=1}^{T}\frac{\mathbf{f}_{\text{text}}(p_{k,t})}{\|\mathbf{f}_{\text{text}}(p_{k,t})\|}$ denotes the mean of the L2-normalized template embeddings of class $c_k$ across $T$ prompt templates $p_{k,1}, \dots, p_{k,T}$ (each template embedding is normalized before averaging, matching the implementation; the cosine in \Cref{eq:clip_pred} then normalizes the averaged vector), with $T=3$ throughout this study. The logit scale of $100$ used in the implementation is omitted because it is a positive constant and does not change the $\arg\max$.

The companion paper~\cite{ma2026agreement} introduced this construct under the name \emph{AnchorProxy}---per-class zero-shot CLIP accuracy under a bbox-cropped input protocol---and used it only as a post-hoc screening signal for pseudo-label reliability. AnchorScore is an independent recomputation of the same construct under the full-frame protocol (\S\ref{sec:input_repr}), the deployment setting in which object bounding boxes are unavailable; this paper provides a systematic empirical evaluation of the construct as an \emph{a priori} predictor of per-class MLLM accuracy.

\subsection{Prompt Design}

To improve robustness without adding computational overhead, text features are averaged over $T=3$ prompt templates for classroom datasets, following the template-averaging strategy of the Smart Classroom Behavior (SCB5) benchmark~\cite{scb_dataset}. Using only the first prompt ($T=1$) yields a consistently lower but still positive correlation at the class level, confirming that multi-prompt averaging adds marginal but consistent benefit:

{\footnotesize
\begin{verbatim}
[ "a photo of a person {cls} in classroom.",
  "a classroom scene showing {cls}.",
  "the action of {cls} in a school environment." ]
\end{verbatim}}

For cross-domain datasets, domain-specific prompts are used (e.g., ``a satellite image of {cls}'' for EuroSAT). For classes with ambiguous names, a manually defined mapping is applied (e.g., the class originally named ``blackBoard'' in the dataset is mapped to the prompt ``standing near a blackboard or whiteboard'') to align with CLIP's semantic space. This mapping has negligible effect: ablating it (using the literal ``blackBoard'' prompt instead) changes AnchorScore from $3.40\%$ to $2.43\%$ ($\Delta=-0.97$~pp; the ablation baseline of $3.40\%$ differs slightly from the canonical $3.16\%$ due to an independent inference run), confirming that the low AnchorScore reflects genuine visual difficulty rather than prompt engineering artifacts.

\textbf{Prompt construction guideline.} The prompt robustness analysis (\Cref{tab:prompt_robustness}) reveals that the prompt \emph{must} include domain context (e.g., ``classroom'' for classroom behavior data). Without domain context, the AnchorScore--MLLM correlation collapses to $\rho\approx0.15$ (e.g., ``a photo of a person raising hand.'' yields $\rho=0.149$, $p=0.63$). Within the constraint that domain context is present, specific phrasing is flexible: four different ``classroom''-inclusive phrasings all yield significant correlations ($\rho=0.60$--$0.77$, all $p<0.05$). Practitioners should construct prompts by appending ``in [domain]'' to the class description, or by using the same template structure as the main experiments.

\subsection{Implementation Details}

The default backbone is ViT-L/14 CLIP~\cite{radford2021clip} pretrained on LAION-2B~\cite{schuhmann2022laion5b}, with FP16 inference and batch size 64. The model has 428M parameters and fits in approximately 17 GB at batch size 64 with FP16. A single forward pass requires approximately $1.1 \times 10^{11}$ FLOPs per image; processing the full SCB5 validation set (5{,}416 images across three subsets) takes approximately 3 minutes. All CLIP inference is performed on a single NVIDIA V100 (32 GB) or A100 (40 GB). Image preprocessing follows CLIP's standard pipeline: resize to 224$\times$224 with bicubic interpolation, normalize with ImageNet statistics. The open-source OpenCLIP library~\cite{cherti2023reproducible} is used for model loading and inference.

MLLM inference uses Ollama-served models (see \Cref{tab:mllms} for the full roster) accessed via the Ollama chat API (\texttt{localhost:11434}) with greedy decoding (\texttt{temperature=0, think=false}), plus LLaVA-1.5-7B-Instruct~\cite{liu2023visual} and Qwen2-VL-7B loaded via HuggingFace Transformers. All model identifiers are the Ollama/HF tag names; see the data availability statement for exact versions.

For the canonical SCB5 evaluation, each MLLM receives the bbox-cropped image regions defined by the benchmark's annotation protocol: each image is cropped to its first annotated person bounding box (read from the dataset's YOLO-format label files), expanded by a 5\% margin of the image dimensions on each side and clamped to the image bounds, together with the classification prompt of \S\ref{sec:mllms}. Frames may contain several person regions, each with its own activity label; the canonical protocol crops the first annotated person region, and each crop carries a single activity label, so the MLLM evaluation is single-label per crop. No detector is run: the crops derive from the dataset's ground-truth annotations. The same crops are used for the bbox-aligned CLIP analysis of \S\ref{sec:input_repr}.

In contrast to CLIP, a 27B-parameter MLLM requires approximately $3 \times 10^{13}$ FLOPs per image (roughly 270$\times$ more), making the full SCB5 validation set run approximately 14 hours on 4 GPUs versus 3 minutes for CLIP on a single GPU. Per GPU on comparable hardware, the 27B MLLM processes roughly $0.03$--$0.07$ images per second versus about 30 for CLIP, so the wall-clock asymmetry does not reduce to a GPU-count artifact.

\section{Experiments}
\label{sec:experiments}

The experiments are organized around four claims, each validated by dedicated analyses.

\textbf{P1 (Existence).} CLIP per-class zero-shot accuracy tracks MLLM annotation difficulty. The primary test is the SCB5 class-level correlation (\S\ref{sec:main_result}); the out-of-sample replication on Stanford40 (\S\ref{sec:stanford40}) is the strongest evidence for this claim.

\textbf{P2 (Specificity).} The signal is not generic to any cheap vision proxy: MLLM self-uncertainty and non-CLIP classifiers (DINOv2, ResNet-50, SigLIP) fail to carry it (\S\ref{sec:extended_baselines}), and its strength varies across CLIP backbones (\S\ref{sec:backbone}).

\textbf{P3 (Shared-difficulty characterization).} The correlation is consistent with a shared class-difficulty factor rather than a CLIP artifact: web-presence and shared-pretraining controls leave the association intact (\S\ref{sec:shared_train}), a consensus analysis is consistent with a predominantly shared component, and a prompt-randomization control shows the association depends on class--prompt alignment, arguing against a pure image-side account (\S\ref{sec:consensus_control}).

\textbf{P4 (Robustness and domain transfer).} The signal persists across input representations (\S\ref{sec:input_repr}) and shows domain-dependent transfer to visually distant domains, attenuating where both models lack competence (\S\ref{sec:cross_domain}).

Every control below rules out a distinct alternative explanation; none is free-standing. The datasets and MLLMs are described first (\S\ref{sec:datasets}--\S\ref{sec:mllms}), then the evidence is presented claim by claim.

\subsection{Datasets}
\label{sec:datasets}

Three classroom behavior sub-datasets from the SCB5 benchmark (Smart Classroom Behavior v5, publicly available on Hugging Face~\cite{scb_dataset}) anchor the evaluation, alongside four cross-domain datasets:

\begin{itemize}
    \item \textbf{TeacherBehavior} (8 classes, 3240 val images): classroom teaching actions including teacher, guide, answer, blackboard-writing, stand, screen, on-stage interaction, and blackboard (standing near a blackboard; ``blackBoard'' in the source dataset, referred to as ``blackboard'' throughout).
    \item \textbf{HandriseReadWrite} (3 classes, 1671 val images): hand-raising, read, and write.
    \item \textbf{BowTurnHead} (2 classes, 505 val images): bowing head and turning head.
    \item \textbf{EuroSAT}~\cite{helber2019eurosat} (10 classes, 27{,}000 images): satellite imagery of land use types.
    \item \textbf{BloodMNIST}~\cite{yang2023medmnist} (8 classes, 3{,}421 images): blood cell microscopy.
    \item \textbf{TissueMNIST}~\cite{yang2023medmnist} (8 classes, 47{,}280 images): kidney tissue microscopy.
    \item \textbf{PathMNIST}~\cite{yang2023medmnist} (9 classes, 7{,}180 images): pathology slides.
\end{itemize}

All datasets used in this study are publicly available benchmarks. The SCB5 sub-datasets are publicly hosted on Hugging Face~\cite{scb_dataset}; the public repository does not currently provide sufficiently detailed documentation regarding the original annotation procedure, participant consent, de-identification process, or an explicit dataset license, so these aspects cannot be independently verified, and the documentation does not state whether participants include minors. This study performs only computational analysis of existing public data, does not redistribute raw images, does not perform face recognition or re-identification, and does not involve direct data collection from human subjects; users should treat the imagery as potentially sensitive and follow their institutional review policies.

\subsection{MLLMs}
\label{sec:mllms}

Six MLLMs from two model families (\Cref{tab:mllms}) are evaluated, with all six tested on TeacherBehavior, HandriseReadWrite, and BowTurnHead. The ``A3B'' and ``A4B'' suffixes denote mixture-of-experts (MoE) architectures with the specified number of active parameters; the Qwen3.5/Qwen3.6 and Gemma-4 releases are documented in their respective official model reports, and the HuggingFace identifiers of the checkpoints evaluated here are listed in \Cref{tab:mllms}. All models are used in a zero-shot setting with a standardized classification prompt. Greedy decoding (temperature=0, max 32 tokens via Ollama API or 15--20 tokens via HuggingFace Transformers) is applied for reproducibility. Output determinism was verified on a subset of 100 images across all models: repeated inference with temperature=0 produced identical outputs in all cases.

\begin{table}[t]
\caption{MLLMs evaluated in this study. HuggingFace model identifiers are provided for reproducibility. The core SCB5 analysis uses the six Qwen and Gemma-4 models (marked $^\dagger$). Coverage of the additional models varies by experiment, as described in \S\ref{sec:mllms}.}
\label{tab:mllms}
\centering
\footnotesize
\setlength{\tabcolsep}{2.5pt}
\begin{tabular}{llc}
\toprule
\textbf{Model} & \textbf{HuggingFace ID} & \textbf{Architecture} \\
\midrule
\multicolumn{3}{l}{\textit{Qwen series}} \\
Qwen3.5-27B$^\dagger$ & Qwen/Qwen3.5-27B & Dense (27B) \\
Qwen3.6-27B$^\dagger$ & Qwen/Qwen3.6-27B & Dense (27B) \\
Qwen3.5-35B-A3B$^\dagger$ & Qwen/Qwen3.5-35B-A3B & MoE (35B, 3B active) \\
Qwen3.6-35B-A3B$^\dagger$ & Qwen/Qwen3.6-35B-A3B & MoE (35B, 3B active) \\
\addlinespace
\multicolumn{3}{l}{\textit{Google Gemma series}} \\
Gemma-4-31B$^\dagger$ & google/gemma-4-31B-it & Dense (31B) \\
Gemma-4-26B-A4B$^\dagger$ & google/gemma-4-26B-A4B-it & MoE (26B, 4B active) \\
\addlinespace
\multicolumn{3}{l}{\textit{Additional MLLMs}} \\
Qwen2-VL-7B & Qwen/Qwen2-VL-7B-Instruct & Dense (7B) \\
LLaVA-1.5-7B & liuhaotian/llava-v1.5-7b-hf & Dense (7B) \\
Qwen2.5-VL-7B & Qwen/Qwen2.5-VL-7B-Instruct & Dense (7B) \\
LLaVA-NeXT-Mistral-7B & llava-hf/llava-v1.6-mistral-7b-hf & Dense (7B) \\
\bottomrule
\end{tabular}
\\[2pt]
\footnotesize{$^\dagger$ Core SCB5 models evaluated on all three sub-datasets (TeacherBehavior, HandriseReadWrite, BowTurnHead).}
\end{table}

The canonical SCB5 evaluation reuses the companion pipeline's instruction format~\cite{ma2026agreement}: the numbered class list is presented with the bbox-cropped image, and the model replies with the class index only. The full-frame image-level pass used for the Mantel analysis (\S\ref{sec:mantel}) instead elicits a category name: ``Classify the classroom activity in this image. Choose exactly one category from the list: [classes]. Output only the category name, nothing else.'' For LLaVA, the prompt is wrapped in a conversational template (\texttt{USER: <image>\ldots ASSISTANT:}); for Qwen2-VL, it is applied via the model's chat template. Full coverage details: TeacherBehavior (8 classes), HandriseReadWrite (3 classes), and BowTurnHead (2 classes) are evaluated on all core models; Qwen2-VL-7B and LLaVA-1.5-7B are additionally evaluated on cross-domain datasets (\S\ref{sec:cross_domain}) and used for SCB5 sensitivity supplementation, while Qwen2.5-VL-7B and LLaVA-NeXT-Mistral-7B are evaluated on cross-domain datasets only.

\subsection{Main Result: AnchorScore--MLLM Correlation}
\label{sec:main_result}

\begin{figure}[t]
\centering
\includegraphics[width=\columnwidth]{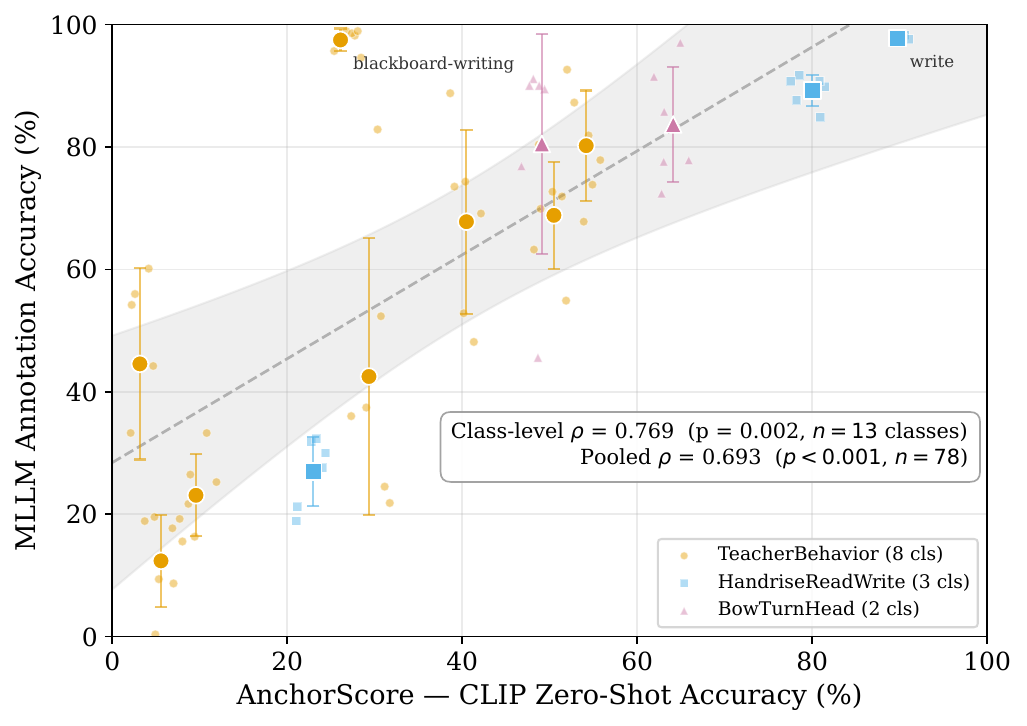}
\caption{AnchorScore vs.\ MLLM accuracy across 78 data points (48 TeacherBehavior, 18 HandriseReadWrite, 12 BowTurnHead). Each point is one MLLM--class pair, jittered horizontally for visibility; colors and marker shapes indicate sub-datasets, and large filled markers show class means with error bars ($\pm1$\,SD across MLLMs). The dashed line and gray band show the OLS fit and its 95\% confidence band on class means; the primary inference is the class-level correlation ($\rho=0.769$, $p=0.002$, $n=13$).}
\label{fig:correlation}
\end{figure}

Does AnchorScore correlate with MLLM per-class accuracy? This is the main empirical question of the paper. The primary test is the class-level Spearman correlation between AnchorScore and mean MLLM accuracy on SCB5 ($n=13$)---the appropriate unit of analysis, because the 78 individual MLLM runs share only 13 distinct AnchorScore values. At this level the correlation is $\rho=0.769$ ($p=0.002$, $n=13$); the wide 95\% bootstrap CI $[0.19, 0.94]$ reflects the limited precision of $n=13$. \Cref{fig:correlation} visualizes the relationship as a strip plot: small translucent markers show individual MLLM runs (jittered horizontally for visibility), whereas large filled markers show class means. The pooled Spearman over all 78 points, shown for reference, is $\rho=0.693$. All subsequent analyses (alternative backbones, non-CLIP baselines, cross-domain validation, ablations) are secondary to this primary class-level test; the Stanford40 experiment (\S\ref{sec:stanford40}) provides independent external replication.

Because the 78 pooled points are not independent (the six MLLMs share 13 AnchorScore values and exhibit substantial pairwise agreement, mean $\rho=0.864$, \S\ref{sec:limitations}), two supplementary checks are fit. A linear mixed-effects model with MLLM-specific random intercepts yields a significant AnchorScore fixed effect ($\beta=0.849$ per pp, 95\% CI $[0.678, 1.019]$, $p<10^{-21}$), but the MLLM-level variance component sits at the boundary of the parameter space (effectively zero), so this model cannot separate between-MLLM variation from residual noise. A class-clustered OLS fit---the design-respecting clustering dimension, since the 78 points share only 13 AnchorScore values (class-level ICC 0.88)---is therefore used: the AnchorScore effect remains significant with cluster-robust standard errors ($\beta=0.849$ per pp, 95\% CI $[0.551, 1.147]$, $p=1.2\times10^{-4}$, $t$ with $G-1=12$ df), confirming that the pooled association is not an artifact of pseudo-replication.

Five sensitivity analyses (\Cref{tab:robustness_all}, Panel~A) confirm the result is not driven by any single class, model, or dataset. The widest leave-one-class-out swing occurs when the largest outlier (``blackboard-writing'', $\Delta=+71.4$~pp) is removed, but every holdout remains positive and significant after Benjamini--Hochberg correction. The TeacherBehavior-only subset remains directionally positive ($\rho=0.595$, $p=0.12$); its non-significance reflects low power ($k=8$ requires $\rho>0.85$ for 80\% power). An independent pooled analysis at $n=23$ (SCB5 + SCB-LLM-202506, a secondary classroom dataset of 10 additional classes from the same repository~\cite{scb_dataset}, $\rho=0.680$, $p<0.001$) reinforces the association with nearly double the sample size. With $n=13$, power reaches 80\% only for $\rho>0.71$. Removing the two-class BowTurnHead subset leaves the association essentially unchanged ($\rho=0.764$, $p=0.006$, $n=11$), indicating that the headline result is not attributable to the structural discreteness of the two-class subset.

\par
\noindent
Per-dataset breakdowns are exploratory and limited by small $k$: TeacherBehavior alone is underpowered ($\rho=0.60$, $p=0.12$, $k=8$), while HandriseReadWrite ($k=3$) and BowTurnHead ($k=2$) are too small for a reliable per-dataset correlation.

Alternative difficulty proxies provide a direct comparison (\Cref{tab:baselines}). Inverse class frequency shows no significant correlation ($\rho=0.187$, $p=0.54$). Among CLIP-derived signals, feature dispersion---the mean pairwise cosine distance among CLIP image embeddings within each class---shows no correlation ($\rho=0.203$, $p=0.51$), mean confidence and margin are moderate but non-significant ($\rho=0.434$, $p=0.14$), and inverse entropy reaches only nominal significance ($\rho=0.604$, $p=0.029$; does not survive Benjamini--Hochberg correction, below). The graded pattern ties the strongest signal to CLIP's text-image \emph{alignment} (whether CLIP maps the class name to the visual concept) rather than the geometric \emph{spread} of image features: a class can be tightly clustered yet misclassified when the prompt does not match the visual content, and such classes are also hard for MLLMs. AnchorScore, which captures alignment directly, is the strongest and most interpretable predictor and is adopted as the primary metric.

Applying Benjamini--Hochberg FDR ($q=0.05$) across the eight alternative predictors in \Cref{tab:baselines} retains AnchorScore as the only significant predictor: inverse entropy reaches nominal significance ($p=0.029$) but does not survive correction (BH threshold $0.006$), and all remaining alternatives are non-significant. Other comparisons in this paper (per-dataset, per-model, cross-domain, LODO, LOMO) are reported as exploratory without multiplicity correction, except the LOCO and Mantel tests, which apply BH-FDR.

\begin{table}[t]
\caption{Baseline comparison at the class level ($n=13$). AnchorScore substantially outperforms all alternative predictors. CLIP inverse entropy reaches nominal significance ($p=0.029$) but does not survive Benjamini--Hochberg correction (threshold $p<0.006$ for 8 predictors, $q=0.05$). \textbf{Bold} = best predictor.}
\label{tab:baselines}
\centering
\footnotesize
\begin{tabular}{lccc}
\toprule
\textbf{Predictor} & \textbf{Spearman} $\rho$ & $p$ & \textbf{Requires} \\
\midrule
\textbf{AnchorScore (CLIP zero-shot acc.)} & \textbf{0.769} & \textbf{0.002} & CLIP only \\
CLIP inverse entropy$^\dagger$ & 0.604 & $0.029^\S$ & CLIP only \\
CLIP confidence / margin$^\dagger$ & 0.434 & 0.14 & CLIP only \\
\midrule
SigLIP zero-shot accuracy & 0.201 & 0.51 & SigLIP model \\
DINOv2 kNN accuracy & +0.050 & 0.87 & DINOv2 + labels \\
ResNet-50 accuracy (supervised on SCB5) & +0.081 & 0.79 & Labels + GPU training \\
\midrule
MLLM self-uncertainty (verbalized) & $-$0.072 & 0.82 & Full MLLM inference \\
\midrule
Feature dispersion$^\ddagger$ & 0.203 & 0.51 & CLIP features \\
Inverse class frequency ($1/N$) & 0.187 & 0.54 & None \\
Random & $\approx 0$ & --- & --- \\
\bottomrule
\end{tabular}
\\[2pt]
\footnotesize{$^\dagger$ CLIP-derived metrics from a separate inference pass. $^\ddagger$ Mean pairwise cosine distance among CLIP image embeddings within each class. $^\S$ Nominal significance only; does not survive Benjamini--Hochberg correction (threshold $p<0.006$, $q=0.05$, 8 predictors). All $p$-values from Spearman rank correlation, two-sided, $n=13$. \textbf{Bold}=best predictor.}
\end{table}
The per-class Deltas (see \Cref{tab:per_class_tb} in Appendix~\ref{app:materials}) reveal a systematic pattern: the largest AnchorScore--MLLM divergences occurred for composite action classes (blackboard-writing +71.4pp, blackboard +41.4pp), whereas simple noun classes showed smaller divergences. The most extreme case, blackboard-writing, achieved 26.1\% by CLIP but 97.5\% by MLLMs. Classes with visually subtle distinctions (e.g., answer at CLIP 5.6\% vs.\ MLLM 12.4\%) showed small absolute Deltas but low accuracy for both models, indicating shared visual difficulty.

MLLM standard deviation also varied substantially across classes. Guide showed the largest inter-model variance ($\pm$22.6~pp, range 21.8

\subsection{Alternative Difficulty Predictors}
\label{sec:extended_baselines}

Is the correlation specific to AnchorScore, or can cheaper alternatives (the MLLM's own verbalized confidence and non-CLIP vision models) capture the same signal? Two critical alternative indicators merit direct comparison: (a)~the MLLM's own verbalized uncertainty, which is a byproduct of inference and requires no additional model, and (b)~a non-CLIP visual classifier, which tests whether the correlation is specific to anchor-class image classifiers in general or merely reflects a generic vision-difficulty shared with any visual model.

\textbf{MLLM self-uncertainty.} For five MLLMs---Qwen3.5-27B, Qwen3.6-27B, Qwen3.6-35B-A3B, Gemma4-26B, and LLaVA-1.5-7B---inference is re-run on SCB5 (30 images/class) asking for a 1--5 confidence rating after classification. At the class level, mean verbalized confidence shows essentially \emph{zero} correlation with MLLM accuracy ($\rho=-0.07$, $p=0.82$, $n=13$). At the pooled level, a weak positive association emerges ($\rho=+0.370$, $p=0.002$, $n=65$; survives Benjamini-Hochberg correction at $q=0.05$), but per-model results reveal high variability---some models are moderately calibrated ($\rho=+0.57$), while others are anti-correlated ($\rho=-0.21$) or produce constant ratings. AnchorScore on this matched subset achieves $\rho=+0.593$ ($p=0.033$; does not survive correction), confirming that MLLM self-uncertainty cannot serve as a reliable class-level diagnostic signal.

\textbf{Non-CLIP visual classifiers.} Three non-CLIP classifiers are evaluated: DINOv2 ViT-B/14~\cite{oquab2024dinov2} via $k$-NN ($k=5$), a ResNet-50~\cite{he2016deep} finetuned on SCB5, and SigLIP~\cite{zhai2023sigmoid} (ViT-SO400M-14, sigmoid pairwise loss). None yields a significant class-level correlation with MLLM accuracy: DINOv2 $\rho=+0.050$ ($p=0.87$), ResNet-50 $\rho=+0.081$ ($p=0.79$), SigLIP $\rho=0.201$ ($p=0.51$; \Cref{tab:baselines}). The SigLIP result is suggestive but confounded by architecture and data differences, so it is framed as directional, not definitive; combined with the consensus control in \S\ref{sec:consensus_control}, it is read as a difference in proxy quality across model families rather than evidence about a specific objective. Baseline models like BLIP-2~\cite{li2023blip2} (zero-shot image-to-text matching with the same prompt templates) and ResNet-50 further compress per-class accuracies into a narrow floor (69\% of classes below 15\%), which itself precludes rank discrimination.

\textbf{MLLM pilot.} The most direct engineering alternative to AnchorScore is a small-MLLM pilot: annotate the same labeled images with a cheap MLLM and use its per-class pilot accuracy as the difficulty ranker. At a matched budget (30 images per class, full-frame), the 7B pilot (LLaVA-1.5-7B, the only evaluated model outside the canonical ground-truth mean) carries no class-difficulty signal ($\rho=-0.03$, $p=0.91$, $n=13$), whereas AnchorScore on the same labeled images attains $\rho=0.778\pm0.040$ (20 subsampling seeds). Pilots from the canonical 26--35B models do rank difficulty ($\rho=0.56$--$0.79$), but they are circular---each pilot model contributes to the ground-truth mean---and cost roughly an hour of 27B inference versus minutes of CLIP inference on the same images. Nor is a non-CLIP encoder a viable pilot substitute: SigLIP, DINOv2, and ResNet-50 show no significant class-level correlation even with the full validation set (\Cref{tab:baselines}), and subsampled pilots of those encoders would be strictly weaker.

\subsection{Robustness to Shared-Training Confounds}
\label{sec:shared_train}

A fundamental concern is whether the AnchorScore--MLLM correlation is a trivial consequence of ``web-presence'': both CLIP and MLLMs are trained on web-scraped data, so classes with higher-frequency names on the web may be better represented in both models' training distributions, yielding a spurious correlation instead of shared visual competence.

Three converging arguments address this concern. First, the web-presence hypothesis is tested directly by controlling for class-name word frequency (log10 frequency from wordfreq~\cite{wordfreq}, based on Common Crawl) as a proxy. Using the pooled 47-class dataset, the Spearman partial correlation between AnchorScore and MLLM accuracy controlling for log word frequency is $\rho=0.594$ ($p=1.1\times10^{-5}$), close to the unadjusted $\rho=0.630$. Word frequency itself shows a weak but significant correlation with both AnchorScore ($\rho=0.310$, $p=0.034$) and MLLM accuracy ($\rho=0.300$, $p=0.040$); the persistence of the AnchorScore--MLLM association after controlling for word frequency indicates that web presence accounts for at most a modest share of the effect rather than explaining it away.

Second, if the correlation were purely a data-overlap artifact, \emph{any} CLIP-derived metric reflecting how readily a class is processed should correlate with MLLM accuracy. Instead, the graded pattern of \S\ref{sec:main_result}---strongest for alignment-based metrics, zero for feature dispersion---indicates that metrics tied to CLIP's discrete class-label alignment carry the shared signal, whereas metrics reflecting only representation geometry do not. The pattern fits a shared-difficulty account (H2/H4; formalized in \S\ref{sec:mechanisms}) but does not fully rule out a data-overlap contribution, since entropy and confidence also plausibly scale with training-data exposure. The word-frequency control and the medical-domain attenuation (below) therefore remain the primary evidence against a pure shared-data explanation.

Third, the BiomedCLIP control~\cite{zhang2023biomedclip} shows that the correlation vanishes on medical data under a domain-specialized encoder: replacing the general CLIP backbone with BiomedCLIP (PubMedBERT-paired, ViT-B/16) on the 25 medical classes yields no class-level correlation with MLLM accuracy ($\rho=0.046$, $p=0.83$), and the general-CLIP medical correlation is likewise non-significant ($\rho=0.214$, $p=0.32$, $n=24$). This pattern fits two interpretations equally well: shared incompetence (neither model handles medical data well) and shared confounds (neither model has medical training data). Distinguishing between them would require evaluating AnchorScore on adversarially generated visual concepts with zero web presence; this is left to future work.

\subsection{Robustness Across CLIP Backbones}
\label{sec:backbone}

Could the finding depend on the specific CLIP backbone or pretraining corpus? To verify that the correlation is not an artifact of a single CLIP model or architecture, AnchorScore is recomputed for all 78 data points using five backbones spanning three model families---two ViT-L/14 and one ViT-B/32 (OpenAI, WIT-400M, and LAION-2B pretraining) plus two from the EVA-02 family (EVA02-L/14 and EVA02-B/16, BAAI merged text--image pretraining)---while keeping the MLLM accuracy values fixed (\Cref{tab:robustness_all}, Panel~B). At the pooled level, four of the five backbones produce significant positive correlations ($\rho=0.377$--$0.693$, all $p<0.001$), while the smallest OpenAI ViT-B/32 shows a positive but non-significant trend ($\rho=0.208$, $p=0.067$). At the class level ($n=13$), the LAION-pretrained ViT-L/14 is the only significant one ($\rho=0.769$, $p=0.002$), with the four others directionally consistent but underpowered (EVA02-B/16 $\rho=0.495$, $p=0.086$; OpenAI ViT-L/14 $\rho=0.473$, $p=0.103$; EVA02-L/14 $\rho=0.412$, $p=0.162$; ViT-B/32 $\rho=0.242$, $p=0.426$). The fact that every backbone---across the OpenAI, LAION, and EVA-02 families and two pretraining corpora---produces a positive association argues against the concern that the AnchorScore--MLLM correlation is merely a LAION- or ViT-L/14-specific artifact.

The attenuation under non-LAION pretraining admits two non-exclusive explanations: (i)~the anchoring effect is not exclusive to LAION, as the directional signal survives a complete change of pretraining corpus \emph{and} model family (EVA-02), and (ii)~given the small class count ($n=13$), these estimates are imprecise: the data neither establish significance for the non-LAION backbones nor contradict the positive-direction hypothesis. The pattern is not a clean capacity gradient: the smallest OpenAI ViT-B/32 ($\rho=0.208$, $p=0.067$) is the weakest, but the two EVA-02 backbones (EVA02-B/16 $\rho=0.408$, $p=2\times10^{-4}$; EVA02-L/14 $\rho=0.377$, $p=7\times10^{-4}$) sit in the same moderate band as the larger OpenAI ViT-L/14 ($\rho=0.405$), and EVA02-B/16 (150M) in fact slightly exceeds OpenAI ViT-L/14 (428M)---so backbone scale alone does not order the signal. What the strongest case shares is the LAION-2B pretraining distribution, which plausibly overlaps with MLLM pretraining data (\S\ref{sec:related}). This pattern (LAION L/14 strong; the four OpenAI and EVA-02 variants clustered in a moderate positive band) is more informative than uniformly significant backbones, as it fits AnchorScore's signal depending on CLIP's specific representational and alignment quality, not a generic vision-property shared by all image classifiers. Fang et al.~\cite{fang2024data} showed that data quality drives CLIP zero-shot performance as strongly as model scale, consistent with a pretraining-data contribution to the LAION-versus-OpenAI/EVA-02 gap observed here.

\subsection{Sensitivity to Input Representation}
\label{sec:input_repr}

CLIP received full frames while the MLLM saw bbox-cropped regions (the first annotated person bounding box from each image's YOLO-format label, cropped with a 5\% margin; \S\ref{sec:method})---could this input mismatch account for the correlation? At deployment time, bounding boxes are unavailable, so AnchorScore must operate on the same full frames a practitioner would feed to their annotation pipeline; full-frame inference is therefore the correct evaluation protocol for the intended use case. To verify that the correlation is not an artifact of this input difference, CLIP is re-run on the same bbox-cropped regions used by the MLLM, reading the YOLO label to obtain pixel coordinates and cropping before applying the standard CLIP transform. The MLLM accuracy data is held fixed.

As shown in \Cref{tab:robustness_all} (Panel~C), aligning CLIP to the bbox input attenuates the pooled correlation (78 MLLM--class points) from $\rho=0.693$ to $\rho=0.424$ ($\Delta\rho=0.269$, paired bootstrap 95\% CI $[0.151, 0.396]$; the CI excludes zero). This attenuation of approximately 39\% is not driven by bbox quality: the per-class bbox area ratio shows no association with the accuracy drop (Spearman $\rho=-0.04$, $p=0.90$, $n=13$). For example, classes with similar crop areas---screen (10.1\%) and read (6.4\%)---both show substantial accuracy \emph{gains} under bbox cropping, but of very different magnitudes ($+52.4$~pp vs.\ $+31.1$~pp), confirming that crop size alone does not explain the pattern. The attenuation instead reflects the combined protocol sensitivity of both families: the canonical MLLM accuracies are themselves tied to the bbox protocol (e.g., Qwen3.5-27B reaches 90.4\% on \textit{guide} but falls to 0\% on \textit{blackboard-writing} under full-frame evaluation; \S\ref{sec:mechanisms}), so the full-frame CLIP ranking is compared against MLLM accuracies measured under a different input condition. The correlation therefore quantifies difficulty alignment under the annotation-pipeline protocol rather than a CLIP-exclusive property.

The bbox-aligned $\rho=0.424$ nonetheless remains significantly positive ($p=1.1\times10^{-4}$), with a 95\% CI that excludes zero. This indicates that exact input alignment is not a precondition for AnchorScore's predictive value, although the strength of that value is sensitive to the specific input representation. Moreover, full-frame inference corresponds to the intended deployment setting where object bounding boxes are unavailable at the time of prediction, making the original experimental design ecologically valid. This input mismatch is specific to SCB5's person-centered action classes, where the dataset's YOLO-format person bounding boxes are used for MLLM inference; it does not apply to Stanford40 (\S\ref{sec:stanford40}) or the cross-domain datasets (\S\ref{sec:cross_domain}), where both CLIP and the MLLM operate on identical full-frame inputs throughout.

\subsection{Zero-Shot Difficulty Across Domains}
\label{sec:cross_domain}

\begin{figure}[t]
\centering
\includegraphics[width=\columnwidth]{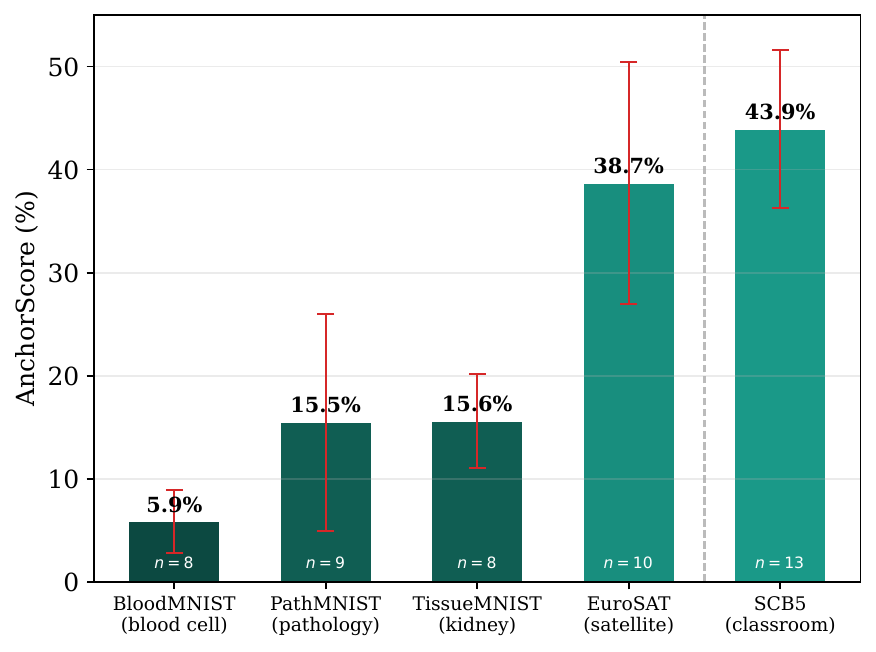}
\caption{AnchorScore across diverse domains spanning satellite imagery (EuroSAT), medical pathology (PathMNIST, BloodMNIST, TissueMNIST), and classroom scenes (SCB5, included as an in-domain reference, separated by a dashed divider). Bars show overall accuracy weighted by class size; error bars show $\pm1$ SEM of per-class accuracies within each dataset. All AnchorScores use CLIP ViT-L/14 (LAION-2B).}
\label{fig:cross_domain}
\end{figure}

Does the AnchorScore association transfer beyond classroom behavior to visually distant domains? AnchorScore is computed on four datasets spanning satellite and medical domains to characterize CLIP's zero-shot capability across visually diverse domains, with cross-domain scores ranging from 5.9\% to 38.7\%. \Cref{fig:cross_domain} visualizes these alongside SCB5 (43.9\%) as an in-domain reference point, showing that classroom scenes and satellite imagery sit at the higher end of CLIP's competence range, while medical domains are substantially harder.

\textbf{Cross-domain MLLM validation.} To examine whether AnchorScore is associated with MLLM difficulty beyond the classroom domain, evaluation is run with \emph{four} MLLMs---Qwen2-VL-7B-Instruct, Qwen2.5-VL-7B-Instruct, LLaVA-1.5-7B, and LLaVA-NeXT-Mistral-7B---on four cross-domain datasets (50 images per class, zero-shot classification). These cross-domain MLLMs are 7B models, substantially smaller than the canonical 26--35B models, so part of any cross-domain attenuation may reflect annotator strength rather than domain. A scale-check with a canonical 27B model (Qwen3.5-27B) on EuroSAT under the same protocol yields 41.8\% mean accuracy---within the 7B range (32.4\%--48.2\%) and again the highest across domains---so the between-domain ordering is robust to annotator scale; the within-dataset signal, by contrast, remains noisy at $n=10$ for every model ($\rho=0.47$ for the 27B versus $-$0.19 to 0.55 across the 7Bs). As shown in \Cref{tab:cross_mllm}, the \emph{between-domain} ordering is partially preserved: EuroSAT (AnchorScore 38.7\%) yields the highest MLLM accuracy (mean 40.3\%), while BloodMNIST (AnchorScore 5.9\%), PathMNIST (AnchorScore 15.5\%), and TissueMNIST (AnchorScore 15.6\%) yield means of 15.4\%, 14.3\%, and 11.9\% respectively. However, the ranking only partly follows AnchorScore: PathMNIST (AnchorScore 15.5\%) shows lower MLLM accuracy than BloodMNIST (AnchorScore 5.9\%), the opposite of what AnchorScore alone would predict. Pooling all per-class data across domains ($n=34$) yields a significant correlation of $\rho=0.462$ ($p=0.006$), indicating that the cross-domain per-class signal strengthens with additional MLLMs.

\begin{table}[t]
\caption{Cross-domain validation: AnchorScore predicts MLLM accuracy beyond classroom domains, though the signal is domain-dependent---the between-domain ordering is partially preserved (EuroSAT highest in every column), while within-dataset correlations are weak on medical data. Four MLLMs evaluated on 50 images per class (zero-shot). Chance rate = $1/C$ per dataset. Pooled per-class Spearman $\rho=0.462$ ($p=0.006$, $n=34$). \textbf{Bold} = best per column.}
\label{tab:cross_mllm}
\centering
\footnotesize
\setlength{\tabcolsep}{3.5pt}
\begin{tabular}{lcccccccc}
\toprule
\textbf{Dataset} & \textbf{Cls} & \textbf{Chance\%} & \textbf{Anchor\%} & \textbf{Qwen2-VL\%} & \textbf{Qwen2.5-VL\%} & \textbf{LLaVA-1.5\%} & \textbf{LLaVA-NeXT\%} & \textbf{Mean\%} \\
\midrule
EuroSAT & 10 & 10.0 & \textbf{38.7} & \textbf{36.8} & \textbf{32.4} & \textbf{48.2} & \textbf{43.6} & \textbf{40.3} \\
TissueMNIST & 8 & 12.5 & 15.6 & 12.3 & 12.5 & 10.5 & 12.5 & 11.9 \\
PathMNIST & 9 & 11.1 & 15.5 & 11.8 & 18.9 & 15.3 & 11.1 & 14.3 \\
BloodMNIST & 8 & 12.5 & 5.9 & 15.5 & 16.0 & 14.8 & 15.3 & 15.4 \\
\bottomrule
\end{tabular}
\end{table}

\subsection{Pooled and Random-Effects Synthesis}
\label{sec:pooled_meta}

\textbf{Pooled class-level analysis.} Given the limited sample size of the classroom-only analysis ($n=13$), all available classes across SCB5 (13 classes) and the four cross-domain datasets (34 classes with matched MLLM data) are pooled for a combined class-level analysis ($n=47$). The pooled Spearman correlation is $\rho=0.630$ ($p<10^{-5}$), confirming a moderate-to-strong positive association between AnchorScore and MLLM accuracy across diverse visual domains. This pooled effect conflates two distinct signals in the aggregation. \emph{Between-domain} variation is present---AnchorScore ranks domain difficulty (EuroSAT 38.7\% $>$ TissueMNIST 15.6\% $>$ PathMNIST 15.5\% $>$ BloodMNIST 5.9\%), whereas MLLM accuracy follows a different ordering (EuroSAT 40.3\% $>$ BloodMNIST 15.4\% $>$ PathMNIST 14.3\% $\approx$ TissueMNIST 11.9\%). \emph{Within-domain} per-class signal, by contrast, shows a mixed pattern: the cross-domain-only class-level correlation is $\rho=0.462$ (bootstrap 95\% CI $[0.15, 0.69]$, $p=0.006$, $n=34$), reaching statistical significance with four MLLMs. The SCB5-only correlation ($\rho=0.769$, $p=0.002$) remains comparable to the cross-domain-only result, and the cross-domain correlation strengthens from $\rho=0.280$ ($p=0.109$, $n=34$) with only two MLLMs to $\rho=0.462$ ($p=0.006$, $n=34$) with four, providing evidence that AnchorScore captures shared visual difficulty beyond the classroom domain, though the signal within individual domains is heterogeneous. Because the pooled class-level test conflates within- and between-domain variation, the between-domain ordering is also tested directly at the domain level: aggregating the 47 classes to $k=7$ domain means yields Spearman $\rho=0.821$ ($p=0.023$), confirming that a substantial share of the pooled association is carried by domain-level difficulty ordering rather than by per-class ranking within domains.

\begin{figure}[t]
\centering
\includegraphics[width=\columnwidth]{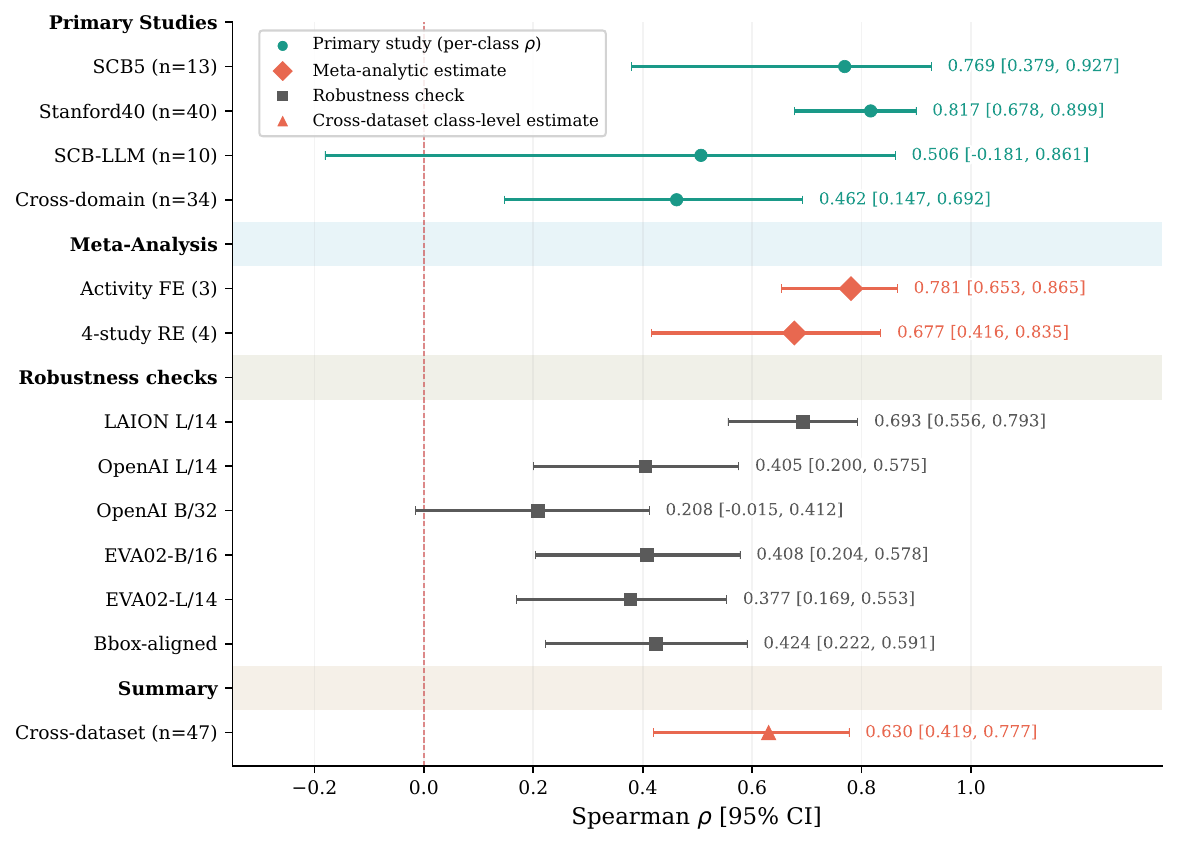}
\caption{Forest plot summarizing the point-estimate robustness checks of the AnchorScore--MLLM association (the range-based sensitivity analyses, e.g., leave-one-class-out and leave-one-MLLM-out, appear in \Cref{tab:robustness_all}). Point estimates with 95\% Fisher $z$-transform CIs (error bars); pooled estimates shown as diamonds. These CIs are methodologically consistent with, but may differ slightly from, the bootstrap CIs in the corresponding tables.}
\label{fig:robustness_forest}
\end{figure}

\Cref{tab:robustness_all} (Panel~D) summarizes a leave-one-dataset-out (LODO) analysis and a random-effects synthesis. Bootstrap confidence intervals accompany the primary analyses; \Cref{fig:robustness_forest} uses Fisher $z$-transform intervals for comparability across analyses of differing sample size. The Stanford40 result (presented in \S\ref{sec:stanford40}; $\rho=0.817$, $n=40$) is similar in magnitude to the SCB5 headline ($\rho=0.769$, $n=13$; Fisher $z=-0.36$, $p=0.72$), whereas Stanford40 differs significantly from the cross-domain estimate ($z=2.66$, $p=0.008$). To summarize: the activity-recognition correlation is consistently strong ($\rho\approx0.78$) across three separately evaluated datasets; the cross-domain signal is directionally consistent but substantially weaker ($\rho=0.462$).

\begin{table*}[t]
\caption{Comprehensive robustness analysis. All sensitivity checks confirm the positive AnchorScore--MLLM association: the signal survives class removal (LOCO), model exclusion (LOMO), backbone substitution (Panel~B; the class-level comparison is underpowered and should not be read as a rigorous capacity gradient), input-representation change (Panel~C), and dataset variation (LODO). \textbf{Bold} = best per panel.}
\label{tab:robustness_all}
\centering
\footnotesize
\setlength{\tabcolsep}{3.5pt}
\renewcommand{\arraystretch}{1.15}
\begin{tabular}{>{\raggedright\arraybackslash}p{2.8cm}>{\raggedright\arraybackslash}p{2.2cm}>{\raggedright\arraybackslash}p{2.8cm}>{\raggedright\arraybackslash}p{5.8cm}}
\toprule
\textbf{Analysis} & \textbf{\boldmath$\rho$\unboldmath / Value} & \textbf{95\% CI / $p$} & \textbf{Key finding} \\
\midrule
\multicolumn{4}{l}{\textbf{A. Sensitivity analyses}} \\
Leave-one-class-out & 0.706--0.909 & all $p<0.02$ & No single class dominates \\
Leave-one-MLLM-out ($n=6$) & 0.725--0.808 & all $p\leq0.005$ & Consistent across MLLMs$^\dagger$ \\
TeacherBehavior-only ($k=8$) & 0.595 & $p=0.12$ & Directional; power-limited \\
\textbf{$+$LLaVA (third family)} & \textbf{0.819} & \textbf{$p<0.001$} & Stronger with broader coverage \\
Mixed-effects (MLLM RE) & $\beta=0.849$ & --- & Mixed-effects sensitivity check; ICC$\approx$0 \\
Class-clustered OLS (13 cls) & $\beta=0.849$ & $[0.551,1.147]$ & Significant at class-level clustering$^\S$ \\
\midrule
\multicolumn{4}{l}{\textbf{B. CLIP backbone variation}} \\
\textbf{ViT-L/14 LAION-2B} & \textbf{0.693 / 0.769} & \textbf{$p=0.002$} & \multirow{5}{5.8cm}{Backbone and pretraining-data variation across three families (LAION, OpenAI, EVA-02): all positive; LAION strongest, no clean capacity gradient} \\
ViT-L/14 OpenAI & 0.405 / 0.473 & $p=0.103$ & \\
ViT-B/32 OpenAI & 0.208 / 0.242 & $p=0.426$ & \\
EVA02-B/16 merged-2B & 0.408 / 0.495 & $p=0.086$ & \\
EVA02-L/14 merged-2B & 0.377 / 0.412 & $p=0.162$ & \\
\midrule
\multicolumn{4}{l}{\textbf{C. Input representation}} \\
Full-frame (original) & 0.693 & $[0.53, 0.811]$, $p<10^{-11}$ & Intended deployment setting \\
Bbox-aligned & 0.424 & $[0.230, 0.583]$, $p=1.1\times10^{-4}$ & Signal survives input change \\
$\Delta$ (full $-$ bbox) & 0.269 & $[0.151, 0.396]$ & --- \\
\midrule
\multicolumn{4}{l}{\textbf{D. Cross-dataset synthesis}} \\
LODO (range, 7 holdouts) & 0.587--0.674 & $p<0.001$ & No dataset drives estimate \\
4-study RE synthesis & 0.677 & $[0.416, 0.835]$, $I^2=62.3\%$ & Significant overall \\
\quad 95\% prediction interval & --- & $[0.126, 0.909]$ & Between-study spread ($\tau^2{=}0.089$) \\
\textbf{Activity-recog subgroup FE} & \textbf{0.781} & \textbf{$[0.653, 0.865]$, $I^2=3.0\%$$^\ddagger$} & Negligible heterogeneity \\
\bottomrule
\end{tabular}
\vspace{2pt}
{\par\footnotesize
Pooled / class-level $\rho$ separated by slash in Panel~B. LODO = leave-one-dataset-out; RE = random-effects; FE = fixed-effect; PI = prediction interval (reflects $\tau^2$ between-study heterogeneity, wider than the RE CI). All pooled estimates use 78 data points (6 MLLMs $\times$ 13 classes) unless noted. \textbf{Bold} = best result per panel. All $p$-values two-sided. $^\dagger$The six MLLMs are highly inter-correlated in per-class rankings (mean pairwise $\rho=0.864$; \S\ref{sec:limitations}), so removing one model provides less independent confirmation than the nominal count implies. $^\ddagger$Based on only three studies; the point estimate should be interpreted cautiously---at $k=3$, $I^2$ confidence intervals are extremely wide. $^\S$HC1 cluster-robust standard errors with $G=13$ class clusters and a $t$ distribution with $G-1=12$ degrees of freedom.
}
\end{table*}

\Cref{fig:robustness_forest} visualizes the point-estimate robustness checks as a forest plot, allowing rapid comparison of effect sizes and confidence intervals across all analyses; the range-based sensitivity analyses (leave-one-class-out, leave-one-MLLM-out) are reported in \Cref{tab:robustness_all}. All point estimates are positive, although several individual confidence intervals include zero (SCB-LLM, and the non-LAION backbones at the class level), consistent with the power analysis in \S\ref{sec:main_result}.

\S\ref{sec:stanford40} tests a domain that shares task structure with classroom behavior while remaining fully independent.

\subsection{Stanford40 Replication}
\label{sec:stanford40}

The strongest replication test comes from an independent, larger-scale human-action dataset. Unlike the cross-domain datasets in \S\ref{sec:cross_domain}, Stanford40 shares classroom behavior's core task structure---recognizing human actions from still images---while being entirely independent in content, providing a stronger test of generalization within the relevant task family.

To further validate the AnchorScore--MLLM association on a substantially larger and independently collected dataset, evaluation is extended to Stanford40 Actions~\cite{yao2011stanford40}, a human action recognition benchmark spanning 40 diverse activity classes (9{,}532 validation images, approximately 238 images per class). The dataset includes everyday actions such as ``applauding,'' ``cutting vegetables,'' ``texting message,'' and ``writing on a book,'' providing a broad test of whether AnchorScore's diagnostic signal generalizes beyond classroom behavior analysis. AnchorScore is computed using the same CLIP ViT-L/14 LAION-2B backbone and three prompt templates (``a photo of a person \{cls\},'' ``a person \{cls\} in a photo,'' ``someone \{cls\}'') that provide appropriate domain context for human action recognition.

Six MLLMs (Qwen3.5-27B, Qwen3.5-35B-A3B, Qwen3.6-27B, Qwen3.6-35B-A3B, Gemma-4-26B-A4B, and Gemma-4-31B) are evaluated on 50 randomly sampled images per class ($2{,}000$ total per MLLM) using zero-shot classification with a category-name prompt on full-frame images. AnchorScore achieves 91.86\% overall accuracy across the 40 classes, with a wide per-class range (47.6\% for ``cutting vegetables'' to 100\% for ``fixing a car'' and ``rowing a boat''). The multi-model class-level Spearman correlation is $\rho=0.817$ (bootstrap 95\% CI $[0.622, 0.930]$, $p<0.001$, $n=40$), nearly identical to the SCB5 headline result of $\rho=0.769$ (Fisher $z=-0.36$, $p=0.72$). All six individual MLLMs show significant positive correlations with AnchorScore ($\rho=0.58$--$0.72$, all $p<0.001$), a notable improvement over SCB5, where the six per-model correlations are individually significant but more variable ($\rho=0.60$--$0.74$, $p=0.004$--$0.029$ at $n=13$) and the two supplemental 7B models do not reach significance. The tighter CI at $n=40$ reflects the substantially larger class sample and provides a more precise estimate of the true effect size. However, 37 of 40 classes exceed 60\% AnchorScore, so the $\rho=0.817$ primarily reflects a few low-AnchorScore classes (e.g., cutting vegetables 47.6\%, texting message 52.9\%) pulling both CLIP and MLLM accuracy downward, not fine-grained ranking across the full difficulty spectrum. On a dataset where nearly all classes are easy for CLIP, AnchorScore's practical value lies in flagging the bottom 3--5 classes for human review, not in full-range ranking.

The Stanford40 result carries two additional implications. First, the dataset covers general human actions, not domain-specific classroom behaviors, demonstrating that AnchorScore's signal is not an artifact of the classroom setting. Second, the effect size ($\rho\approx0.81$) is similar in magnitude to the SCB5 result, and the three activity-recognition datasets (SCB5, Stanford40, SCB-LLM-202506) show negligible heterogeneity ($I^2=3.0\%$) in the synthesis (\S\ref{sec:pooled_meta}), indicating a common underlying association across activity-recognition tasks. This replication, with 40 independent classes and over three times the sample size of the primary analysis, substantially strengthens the evidence that CLIP zero-shot accuracy reliably flags classes where MLLMs will struggle.

\section{Applications}
\label{sec:applications}

\begin{figure}[t]
\centering
\includegraphics[width=\columnwidth]{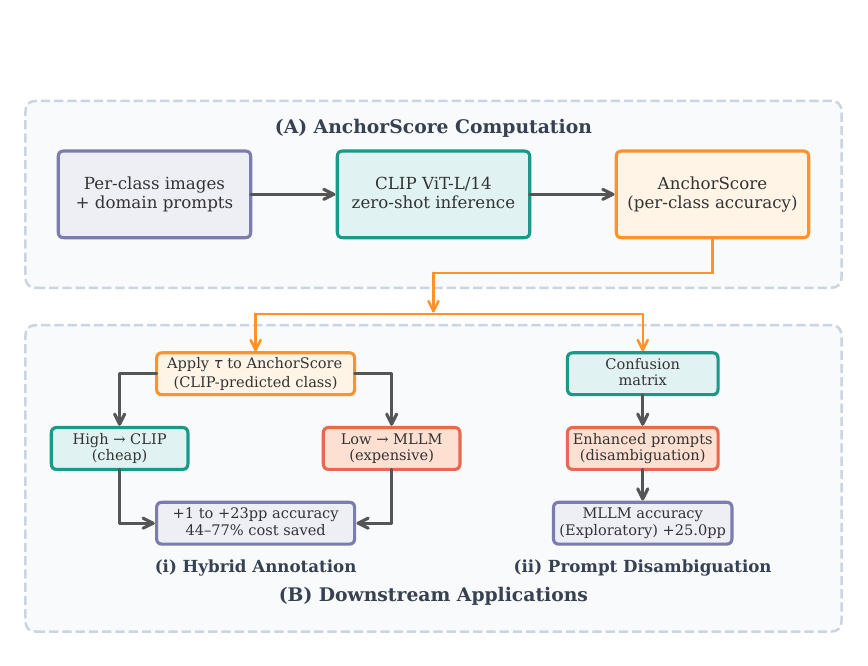}
\caption{The AnchorScore pipeline. (A)~AnchorScore computation: per-class validation images with domain prompts pass through frozen CLIP ViT-L/14 zero-shot inference, yielding a per-class AnchorScore. (B)~Downstream workflows: (i)~hybrid routing thresholds the AnchorScore of CLIP's \emph{predicted} class ($\tau$), accepting high-score predictions from CLIP and sending low-score images to the MLLM (gains of $+$1 to $+$23~pp at 44--77\% cost savings across the deployable datasets; \Cref{tab:hybrid} Panel~A); (ii)~prompt disambiguation derives enhanced MLLM prompts from CLIP's confusion matrix (combined $+$25.0~pp with direct visual-distinction descriptions; \S\ref{sec:prompt_opt}).}
\label{fig:pipeline}
\end{figure}

Beyond its diagnostic value, AnchorScore enables three workflow analyses that bear directly on MLLM annotation pipelines: deployable hybrid routing, prompt-disambiguation experiments, and review-priority ranking (\Cref{fig:pipeline}). All three are demonstrated on the SCB5 dataset using the reference AnchorScore values from \S\ref{sec:main_result}. One methodological note applies to the routing analyses specifically: they combine cached per-image CLIP predictions with the canonical per-class MLLM accuracy rates (per-image MLLM predictions under the canonical bbox protocol are not cached), so the MLLM branch is modeled by its per-class expected accuracy and the reported routing accuracies are expected values; realized runs validating these estimates under both protocols are reported in \S\ref{sec:hybrid}.

\subsection{Hybrid CLIP/MLLM Routing}
\label{sec:hybrid}

AnchorScore's ability to rank per-class difficulty suggests a routing question: can a cheap CLIP classifier handle some images while an MLLM handles the rest? The deployable policy routes each image by CLIP's \emph{predicted} class---no ground-truth label is used at routing time. For an image $x$, if $\text{AnchorScore}(\text{CLIP}(x)) \geq \tau$, CLIP's prediction is accepted; otherwise $x$ is sent to a single fixed MLLM. This abstention-style decision is in the spirit of selective prediction~\cite{geifman2017selective}, but is made from a class-level \emph{a priori} signal (the AnchorScore of CLIP's predicted class) rather than instance-level confidence; the positioning relative to that literature is given in \S\ref{sec:related}. The policy is evaluated on the 5{,}416 SCB5 validation images using the cached per-image CLIP predictions (same protocol as the canonical AnchorScore; per-class agreement within 0.5~pp) and the canonical per-class MLLM accuracies. The MLLM branch contributes its per-class expected accuracy---the same assumption underlying the class-aware bound below, since per-image MLLM predictions under the canonical bbox protocol are not cached. This estimate is therefore an expected value, not a realized pipeline run.

\Cref{tab:hybrid} (Panel~A) reports the deployable results. On TeacherBehavior ($\tau=45$), the pipeline reaches 55.7\% accuracy---$+$23.3~pp over CLIP-only (32.4\%)---while sending 56.5\% of images to the MLLM. Every one of the six MLLMs yields a positive gain ($+$21.8 to $+$24.6~pp), so the result does not depend on MLLM choice. On BowTurnHead ($\tau=50$), the gain is $+$21.6~pp at 60.8\% cost saving. On HandriseReadWrite, no threshold improves over CLIP-only; the deployable optimum there is CLIP alone, which the table reports honestly. The reported $\tau^*$ values are retrospectively selected operating points on the evaluation data; \Cref{sec:discussion} describes a budget-based selection rule for cold-start deployment.

The deployable policy can even exceed the class-aware bound computed from true classes. On TeacherBehavior, predicted-class routing attains 55.7\% versus 44.7\% for true-class routing at the same $\tau$: CLIP's predicted class doubles as a confidence signal, because images on which CLIP predicts a high-AnchorScore class are precisely where CLIP is most reliable (61.9\% correct on the routed subset), while the remaining, harder images go to the stronger MLLM. True-class routing cannot exploit this signal, so it is not an upper bound on predicted-class policies---only on policies that decide purely by the true class. Two matched-cost routing baselines confirm that AnchorScore, not CLIP confidence, drives the gain: at the same 56.5\% MLLM budget on TeacherBehavior, random routing attains 45.0\% ($\pm$0.4\% over 200 seeds) and CLIP max-logit-confidence routing attains 44.4\%, versus 55.7\% for AnchorScore routing. (The confidence-routing curve rises monotonically with the MLLM budget and peaks at the all-MLLM baseline, so its maximum is not an informative comparison.)

The expected-value estimate is validated by realized pipeline runs under both evaluation protocols. Under the benchmark's canonical bbox-crop protocol, the MLLM branch (1{,}832 routed images) executed with Qwen3.5-27B yields a realized overall accuracy of 54.1\%---within 1.8~pp of the Qwen3.5-27B expected value (55.9\%) and still $+$21.7~pp over CLIP-only. Under the deployment condition, where the fallback receives full frames (no annotation-derived bounding box is available at prediction time), the same pipeline yields 49.9\%---$+$17.5~pp over CLIP-only. The routing decision itself uses only CLIP's predicted class and no ground truth; the bounding-box crops enter only the canonical MLLM evaluation, so the 4.2~pp gap between the two realized runs quantifies the protocol difference rather than an information leak, and the realized deployment gain remains substantial. A calibration sample (30 images/class) shows the Ollama-served backend differs from the canonical HuggingFace run by only $-1.1$~pp on average, so the small realized--expected gap reflects the routed subset being slightly harder than the class average rather than a failed estimate.

\Cref{tab:hybrid} (Panel~B) reports the class-aware bound as a reference for the class-known deployment setting (e.g., curriculum-based collection where images are gathered per class). On Stanford40 the bound gains $+$4.7~pp at 83\% cost saving, discussed below as a complementary regime.

\begin{table}[t]
\caption{Hybrid CLIP/MLLM routing. \textbf{Panel A (deployable):} each image is routed by CLIP's \emph{predicted} class; accuracy is an expected value over per-image CLIP predictions and per-class MLLM rates, validated by a realized Qwen3.5-27B run on TeacherBehavior (54.1\% realized vs.\ 55.9\% expected; \S\ref{sec:hybrid}). ``Fixed range'' = per-MLLM gains with a single committed MLLM. \textbf{Panel B (class-aware bound):} images are routed by their \emph{true} class, bounding only policies that decide by true class; useful for class-known settings. Cost saved: Panel A = fraction of images not sent to the MLLM (CLIP cost treated as negligible); Panel B = savings under a 100:1 MLLM:CLIP per-image cost ratio.}
\label{tab:hybrid}
\centering
\footnotesize
\setlength{\tabcolsep}{2.5pt}
\begin{tabular}{lccccc}
\toprule
\multicolumn{6}{l}{\textit{Panel A: deployable predicted-class routing}} \\
\textbf{Dataset} & \textbf{CLIP} & \textbf{Deploy acc} & \textbf{Fixed range} & \textbf{Cost} & $\tau^*$ \\
 & & (mean MLLM) & (gain, pp) & saved & \\
\midrule
TeacherBehavior (8 cls) & 32.4\% & 55.7\% & $+$21.8 to $+$24.6 & 44\% & 45 \\
HandriseReadWrite (3 cls) & 61.0\% & 61.0\% & $+$0.0 (CLIP-only) & 100\% & 0 \\
BowTurnHead (2 cls) & 60.4\% & 82.0\% & $+$19.5 to $+$23.4 & 61\% & 50 \\
Stanford40 (40 cls) & 91.9\% & 93.3\% & $+$1.0 to $+$1.6 & 77\% & 95 \\
\midrule
\multicolumn{6}{l}{\textit{Panel B: class-aware bound (true-class routing, reference)}} \\
\textbf{Dataset} & \textbf{CLIP} & \textbf{Best bound} & \textbf{Fixed range} & \textbf{Cost} & $\tau^*$ \\
 & & (best MLLM) & (gain, pp) & saved & \\
\midrule
TeacherBehavior (8 cls) & 32.6\% & 43.6\% & $+$8.7 to $+$16.9 & 53\% & 45 \\
HandriseReadWrite (3 cls) & 60.9\% & 64.1\% & $-$1.5 to $+$3.4 & 63\% & 45 \\
BowTurnHead (2 cls) & 60.8\% & 66.9\% & $-$0.8 to $+$9.3 & 77\% & 55 \\
Stanford40 (40 cls) & 91.9\% & 96.5\% & $+$3.8 to $+$4.7 & 83\% & 80 \\
\bottomrule
\end{tabular}
\\[2pt]
\footnotesize{Panel A uses the per-image CLIP baseline (32.4\%, 61.0\%, 60.4\%, 91.9\%); Panel B uses the canonical weighted AnchorScore baseline (32.6\%, 60.9\%, 60.8\%, 91.9\%), a $\leq$0.2~pp difference. Versus the best all-MLLM pipeline, deployable routing incurs penalties of 2.6~pp (TeacherBehavior), 9.9~pp (HandriseReadWrite), 4.5~pp (BowTurnHead), and 4.7~pp (Stanford40) at 44--77\% cost reduction---a favorable cost--accuracy trade-off on TeacherBehavior and BowTurnHead, none on HandriseReadWrite, and a modest one on Stanford40. The class-aware bound (Panel~B) incurs penalties of 16.0~pp (TeacherBehavior), 6.8~pp (HandriseReadWrite), 21.3~pp (BowTurnHead), and 1.5~pp (Stanford40).}
\end{table}

\begin{figure}[t]
\centering
\includegraphics[width=\columnwidth]{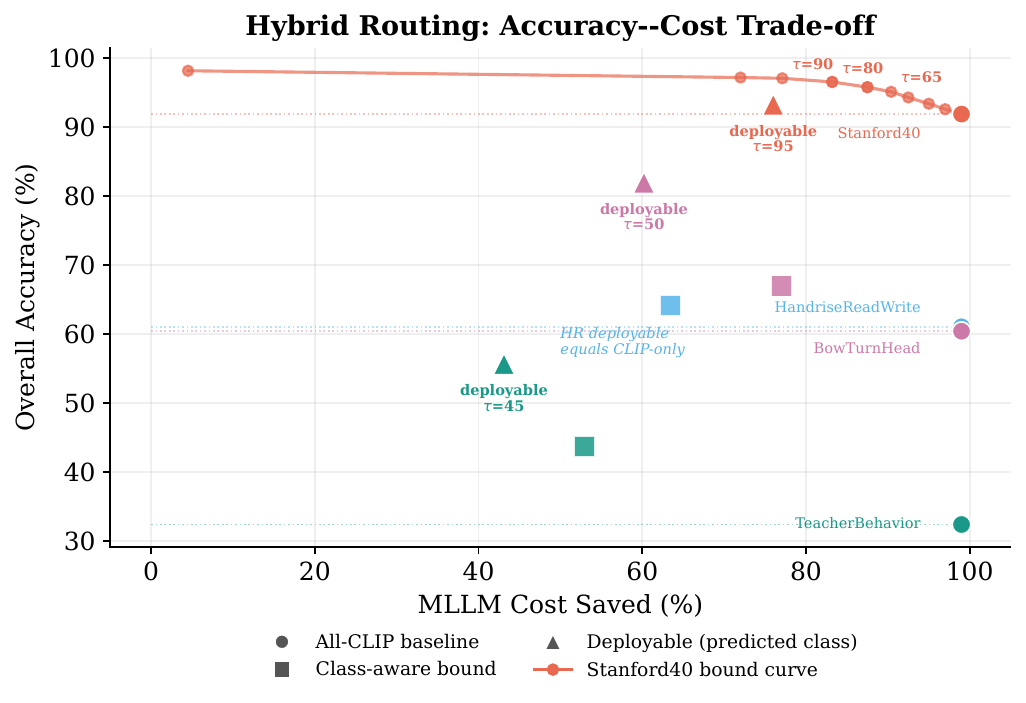}
\caption{Pareto visualization of hybrid routing: overall accuracy ($y$-axis) vs.\ MLLM cost saved ($x$-axis); better operating points lie toward the upper-left. \textbf{Open circles} with dashed horizontal lines are the all-CLIP baselines (no MLLM calls, 99\% cost saving): TeacherBehavior 32.4\%, HandriseReadWrite 61.0\%, BowTurnHead 60.4\%, Stanford40 91.9\%. \textbf{Filled squares} are the class-aware bound (routing decided on the \emph{true} class; a theoretical reference, not deployable). \textbf{Filled triangles} are the deployable predicted-class operating points: TeacherBehavior 55.7\% at 43\% saving ($+$23.3~pp), BowTurnHead 82.0\% at 60\% ($+$21.6~pp), Stanford40 93.3\% at 76\% ($+$1.4~pp); a triangle above its baseline means accuracy gained while cost is saved. HandriseReadWrite's deployable optimum is CLIP-only, so no triangle is drawn. The \textbf{Stanford40 curve} (connected circles) shows the full bound trade-off as the threshold $\tau$ varies from 0 to 100; only Stanford40 is drawn as a curve because its 40-class bound is smooth, whereas the SCB5 bounds (8/3/2 classes) are step functions that would clutter the figure. The $x$-axis uses the 100:1 MLLM:CLIP per-image cost-ratio convention (all-CLIP $=$ 99\%); the deployable markers differ from \Cref{tab:hybrid} Panel~A's image-fraction savings by $\leq$0.8~pp.}
\label{fig:pareto}
\end{figure}

\noindent\textbf{Stanford40: a complementary regime.} Stanford40 illustrates the opposite end of the difficulty spectrum: CLIP already achieves 91.9\% accuracy, and 37 of 40 classes exceed 60\% AnchorScore. The deployable routing at $\tau=95$ gains $+$1.4~pp (from 91.87\% to 93.26\%) while saving 76.8\% of MLLM cost, with a narrow per-model range ($+$1.0 to $+$1.6~pp). The class-aware bound is larger---$+$4.7~pp (from 91.86\% to 96.53\%) at 83\% saving---because true-class routing can target the seven classes below $\tau=80$ exactly, whereas predicted-class routing must tolerate CLIP's own misclassifications. Among the bound's routed classes, the three lowest---cutting vegetables (47.6\%), texting message (52.9\%), and writing on a book (58.9\%)---see their accuracy jump from 48--59\% to 84--94\%. Together, the SCB5 and Stanford40 results show that routing adapts to the dataset's difficulty profile: it delivers large accuracy gains when CLIP is weak (TeacherBehavior, BowTurnHead), a modest gain when CLIP is already strong (Stanford40), and none where the class spread is uninformative (HandriseReadWrite). \Cref{fig:pareto} visualizes this accuracy--cost trade-off.

\subsection{AnchorScore-Guided Prompt Disambiguation (Exploratory)}
\label{sec:prompt_opt}

Beyond ranking difficulty, CLIP's confusion patterns could diagnose \emph{why} a class is hard and guide targeted MLLM prompt improvements. The per-image confusion matrix identifies which class pairs are visually confusable. Because the Mantel analysis in \S\ref{sec:mantel} predicts that CLIP's confusion \emph{structure} will not transfer systematically to the MLLM, these confusable pairs are treated as \emph{hypotheses}---which visual distinction to clarify---whose cross-model transfer must be verified per model rather than assumed. For each class with low AnchorScore, the most confused class was identified from this matrix, and a disambiguation hint was injected into the MLLM prompt, replacing the standard instruction with a version that names the likely confusion pair. Two prompt strategies are evaluated: a negation-style hint (``this image may show \{cls\}, not \{confuser\}'') and a direct description of the visual distinction between the two classes. \Cref{tab:prompt_opt_negation,tab:prompt_opt_direct} summarize the results across five low-AnchorScore classes and two MLLMs. This section is \emph{exploratory}: the direct-description strategy was adopted after observing the negation results on the same images, and all per-pair significance statements are reported without multiplicity correction, consistent with this framing. To test whether the direct gains reflect CLIP's confuser \emph{selection} rather than the mere addition of guidance text, we ran two controlled conditions on the identical 20 images per class (temperature 0, so all conditions are paired per image): a \emph{generic} condition with the same skeleton plus a length-matched ``categories can look similar; look carefully'' note naming no pair, and a \emph{wrong-pair} condition using the direct template with the \emph{second}-most-confused CLIP confuser substituted (identical sentence structure and length; BowHead, a two-class dataset, receives only the generic control). \Cref{tab:prompt_opt_confuser} reports both tests. The generic control shows the gains are not an artifact of added prompt text: on blackBoard and answer---the two lowest-AnchorScore classes---generic guidance yields 0--5\% accuracy while the direct condition reaches 55--80\% ($\Delta$ direct$-$generic significant in 4 of 10 class--model pairs, all on these two classes). The wrong-pair control shows the effect attaches to the CLIP-derived pair specifically: direct beats wrong-pair significantly in 3 of 8 comparisons (answer on both models, $+$60.0/$+$55.0~pp; blackBoard on Qwen3.6-27B, $+$55.0~pp), with no comparison in which wrong-pair significantly beats direct. On stand, read, and BowHead no condition differs significantly, consistent with the direct condition's own pattern. Read together with the Mantel null (\S\ref{sec:mantel}), the controls support a narrow reading: where clarification helps at all, it helps because the \emph{specific} CLIP-identified confuser is named---concentrated on the classes with the lowest AnchorScore---rather than from generic guidance or from any confusable pair.

\begin{table}[t]
\caption{Prompt disambiguation with negation-style hints (``this image may show \{cls\}, not \{confuser\}.''). Gains are limited (combined $+$11.0~pp) and can backfire (blackboard on Qwen3.6-27B: $-$30.0~pp).}
\label{tab:prompt_opt_negation}
\centering
\footnotesize
\setlength{\tabcolsep}{3pt}
\begin{tabular}{lccccccc}
\toprule
& & \multicolumn{3}{c}{\textbf{Qwen3.5-27B}} & \multicolumn{3}{c}{\textbf{Qwen3.6-27B}} \\
\cmidrule(lr){3-5} \cmidrule(lr){6-8}
\textbf{Class} & \textbf{Anchor} & Std & Neg & $\Delta$ & Std & Neg & $\Delta$ \\
\midrule
blackboard & 3.2\% & 5.0 & 10.0 & $+$5.0$^\dagger$ [0,~15] & 35.0 & 5.0 & $-$30.0$^*$ [$-$50,~$-$10] \\
answer & 5.6\% & 10.0 & 40.0 & $+$30.0$^*$ [10,~50] & 10.0 & 15.0 & $+$5.0$^\dagger$ [0,~15] \\
stand & 9.6\% & 25.0 & 70.0 & $+$45.0$^*$ [25,~65] & 15.0 & 20.0 & $+$5.0$^\dagger$ [$-$10,~20] \\
read & 23.0\% & 65.0 & 80.0 & $+$15.0$^\dagger$ [0,~30] & 65.0 & 80.0 & $+$15.0$^\dagger$ [0,~30] \\
BowHead & 49.1\% & 80.0 & 85.0 & $+$5.0$^\dagger$ [$-$10,~20] & 75.0 & 90.0 & $+$15.0$^\dagger$ [0,~30] \\
\textbf{Average} & --- & 37.0 & 57.0 & \textbf{\boldmath$+$\unboldmath20.0} & 40.0 & 42.0 & \textbf{\boldmath$+$\unboldmath2.0} \\
\bottomrule
\end{tabular}
\\[2pt]
\footnotesize{All accuracy values in \%. $n=20$ images/class per condition, identical images across conditions. $\Delta$ vs.\ the standard prompt; 95\% CI in brackets from $B=20{,}000$ paired bootstrap resamples; $^*$CI excludes zero, $^\dagger$CI includes zero. Standard-prompt baselines (Std) are the same as in \Cref{tab:prompt_opt_direct} because identical images were used across all conditions.}
\end{table}

\begin{table}[t]
\caption{Prompt disambiguation with direct visual-distinction descriptions. Substantially more effective than negation (combined $+$25.0~pp vs.\ $+$11.0~pp), with 3/10 class--model pairs significantly improved and no significant decreases.}
\label{tab:prompt_opt_direct}
\centering
\footnotesize
\setlength{\tabcolsep}{3pt}
\begin{tabular}{lccccccc}
\toprule
& & \multicolumn{3}{c}{\textbf{Qwen3.5-27B}} & \multicolumn{3}{c}{\textbf{Qwen3.6-27B}} \\
\cmidrule(lr){3-5} \cmidrule(lr){6-8}
\textbf{Class} & \textbf{Anchor} & Std & Dir & $\Delta$ & Std & Dir & $\Delta$ \\
\midrule
blackboard & 3.2\% & 5.0 & 55.0 & $+$50.0$^*$ [30,~70] & 35.0 & 65.0 & $+$30.0$^\dagger$ [$-$10,~65] \\
answer & 5.6\% & 10.0 & 80.0 & $+$70.0$^*$ [50,~90] & 10.0 & 65.0 & $+$55.0$^*$ [35,~75] \\
stand & 9.6\% & 25.0 & 20.0 & $-$5.0$^\dagger$ [$-$25,~15] & 15.0 & 20.0 & $+$5.0$^\dagger$ [0,~15] \\
read & 23.0\% & 65.0 & 75.0 & $+$10.0$^\dagger$ [0,~25] & 65.0 & 75.0 & $+$10.0$^\dagger$ [0,~25] \\
BowHead & 49.1\% & 80.0 & 90.0 & $+$10.0$^\dagger$ [$-$10,~30] & 75.0 & 90.0 & $+$15.0$^\dagger$ [0,~30] \\
\textbf{Average} & --- & 37.0 & 64.0 & \textbf{\boldmath$+$\unboldmath27.0} & 40.0 & 63.0 & \textbf{\boldmath$+$\unboldmath23.0} \\
\bottomrule
\end{tabular}
\\[2pt]
\footnotesize{All accuracy values in \%; protocol and notation as in \Cref{tab:prompt_opt_negation}. Standard-prompt baselines (Std) are the same as in \Cref{tab:prompt_opt_negation} because identical images were used across all conditions.}
\end{table}

\begin{table}[t]
\caption{Confuser controls for the direct condition, on the identical 20 images per class. Gen: generic note naming no pair; Wr: direct template with the second-most-confused CLIP confuser. $\Delta$ = direct $-$ Wr; direct accuracies are in \Cref{tab:prompt_opt_direct}.}
\label{tab:prompt_opt_confuser}
\centering
\footnotesize
\setlength{\tabcolsep}{3pt}
\begin{tabular}{lccccccc}
\toprule
& & \multicolumn{3}{c}{\textbf{Qwen3.5-27B}} & \multicolumn{3}{c}{\textbf{Qwen3.6-27B}} \\
\cmidrule(lr){3-5} \cmidrule(lr){6-8}
\textbf{Class} & \textbf{Anchor} & Gen & Wr & $\Delta$ & Gen & Wr & $\Delta$ \\
\midrule
blackboard & 3.2\% & 0.0 & 25.0 & $+$30.0$^\dagger$ [0,~55] & 5.0 & 10.0 & $+$55.0$^*$ [35,~75] \\
answer & 5.6\% & 0.0 & 20.0 & $+$60.0$^*$ [40,~80] & 0.0 & 10.0 & $+$55.0$^*$ [35,~75] \\
stand & 9.6\% & 20.0 & 45.0 & $-$25.0$^\dagger$ [$-$50,~0] & 10.0 & 20.0 & $+$0.0$^\dagger$ [0,~0] \\
read & 23.0\% & 70.0 & 75.0 & $+$0.0$^\dagger$ [$-$15,~15] & 70.0 & 75.0 & $+$0.0$^\dagger$ [0,~0] \\
BowHead & 49.1\% & 80.0 & --- & --- & 85.0 & --- & --- \\
\bottomrule
\end{tabular}
\\[2pt]
\footnotesize{Gen/Wr are accuracy in \%; $\Delta$ = direct $-$ Wr, 95\% CI from $B=20{,}000$ paired bootstrap; $^*$CI excludes zero, $^\dagger$CI includes zero. $\Delta$ direct$-$generic is significant in 4/10 (blackBoard $+$55.0/$+$60.0, answer $+$80.0/$+$65.0, both models) and $\Delta$ direct$-$Wr in 3/8, all direct-favoring; no significant differences on stand, read, or BowHead. BowHead has no wrong pair (two-class dataset).}
\end{table}

The negation-style hints (\Cref{tab:prompt_opt_negation}) improved only two of ten class--MLLM pairs significantly (answer and stand on Qwen3.5-27B, $+$30.0 and $+$45.0~pp), while Qwen3.6-27B on blackboard significantly decreased ($-$30.0~pp, 95\% CI [$-$50.0, $-$10.0]), confirming that label-negation prompts can backfire when the confusion pattern differs across models.

The alternative suggested by these results was therefore tested: prompts that describe the visual distinction directly, without negating the confused label (\Cref{tab:prompt_opt_direct}). The direct descriptions improved all but one class--model pair, with the largest gains concentrated in the two lowest-AnchorScore classes (blackboard and answer, $+$50.0 and $+$70.0~pp on Qwen3.5-27B; answer $+$55.0~pp on Qwen3.6-27B). The most striking reversal is blackboard on Qwen3.6-27B: the negation hint reduced accuracy by 30~pp while the direct description increased it by 30~pp.

Direct description significantly changed the outcome relative to negation on five of ten pairs (better on four, including both blackboard and answer on both models, $+$40 to $+$60~pp; worse on one: stand on Qwen3.5-27B, $-$50~pp relative to negation). These results indicate that CLIP's confusion pattern provides useful hypotheses about \emph{which} visual distinction to clarify, and that describing that distinction directly is a safer and more effective strategy than negating the confused label, though gains remain model-specific.

\subsection{Review-Priority Prediction}
\label{sec:ranking}

An aggressive prescriptive use of AnchorScore---shifting the training-sample budget toward low-AnchorScore classes---was first tested and is not supported by the experiments; the supported prescriptive role, evaluated below, is review-priority ranking. In the rejected budget-shifting form, a linear-probe simulation on frozen CLIP features (20\% training budget, 5 seeds) showed that neither a capped budget shift ($\pm 30\%$, constrained to $[0.5\times,\,2.0\times]$ uniform) nor naive $1/$AnchorScore weighting beats uniform sampling: the capped strategy improved accuracy on only one of three datasets (HandriseReadWrite $+$1.4~pp) and degraded the other two (TeacherBehavior $-$0.6~pp, BowTurnHead $-$11.3~pp), while naive weighting over-allocated budget to the most extreme low-score class (blackboard at 3.2\% received the largest share) and underperformed uniform sampling on average.

This is formalized as follows. For each of the three SCB5 datasets, the ``hard'' classes are defined as those with MLLM accuracy below the dataset median (i.e., the $k/2$ hardest classes out of $k$ total). AnchorScore ranks all classes from lowest to highest, and the measure of interest is whether the lowest-AnchorScore classes capture these hard classes. Two metrics are reported: (i)~Area Under the ROC Curve (AUC), where a random ranker achieves 0.5 and perfect ranking achieves 1.0; and (ii)~hit rate at $k/3$ (precision and recall when reviewing the third of classes with the lowest AnchorScore). Results are summarized in \Cref{tab:ranking}.

\begin{table}[t]
\caption{Review priority ranking: AnchorScore identifies hard classes (below-median MLLM accuracy) with AUC$>$0.84 on both TeacherBehavior and Stanford40, supporting review-priority ranking. Bootstrap 95\% CI (5000 replicates) for TeacherBehavior and Stanford40; HandriseReadWrite ($k=3$) and BowTurnHead ($k=2$) have trivially perfect AUC at such small $k$.}
\label{tab:ranking}
\centering
\footnotesize
\begin{tabular}{lcccc}
\toprule
\textbf{Dataset} & $k$ & \textbf{AUC} & \textbf{Precision@$k/3$} & \textbf{Recall@$k/3$} \\
\midrule
\textbf{TeacherBehavior} & 8 & \textbf{0.938} [0.667, 1.000] & \textbf{1.000} & \textbf{0.750} \\
\textbf{Stanford40} & 40 & \textbf{0.842} [0.690, 0.958] & 0.857 & 0.632 \\
HandriseReadWrite$^\dagger$ & 3 & 1.000 & 1.000 & 1.000 \\
BowTurnHead$^\dagger$ & 2 & 1.000 & 1.000 & 1.000 \\
\bottomrule
\end{tabular}
\\[4pt]
\footnotesize{$^\dagger$ Datasets with $k<5$ produce trivially perfect AUC values due to small class count; only TeacherBehavior ($k=8$) and Stanford40 ($k=40$) provide meaningful ranking evaluations. \textbf{Bold}=best validated results. All ranking metrics use median-MLLM-accuracy threshold per dataset.}
\end{table}

AnchorScore achieves high AUC across SCB5: on TeacherBehavior ($k=8$), AUC=0.938 (95\% CI $[0.667, 1.000]$); a permutation test rejects the null ($p=0.028$, though not surviving BH correction), and reviewing the lowest-anchor $k/3=3$ classes catches 3 of 4 hard classes. On Stanford40 ($k=40$), AUC=0.842 (95\% CI $[0.690, 0.958]$, $B=5{,}000$); the permutation test is conclusive ($p<0.001$, surviving BH correction), and reviewing $k/3=14$ classes catches 12 of 19 hard classes (precision $0.857$, recall $0.632$, $1.8\times$ random). HandriseReadWrite ($k=3$) and BowTurnHead ($k=2$) produce trivially perfect AUC due to small class counts.
Unlike the sampling experiment reported above, the ranking framing directly leverages AnchorScore's strength as a relative difficulty comparator. The Stanford40 evaluation provides the robust validation at $k=40$, while the TeacherBehavior result ($k=8$) is directionally consistent but imprecise. Practitioners with limited review capacity should note that meaningful ranking evaluation requires more than the 2--3 classes available in the smallest SCB5 subsets.

\section{Discussion}
\label{sec:discussion}

\subsection{Mechanisms and the Limits of Shared Structure}
\label{sec:consensus_control}
\label{sec:mechanisms}

What explains the AnchorScore--MLLM association? Four competing (non-exclusive) hypotheses are considered, a cross-model consensus control and a prompt-randomization control are reported that constrain the mechanism, and then a stronger structural prediction implied by the most plausible account is tested.

\textbf{Cross-model consensus control.} If the association reflected a CLIP-\emph{specific} capability, AnchorScore would explain per-class accuracy beyond what other MLLMs already reveal. This is tested by conditioning the pooled 78-point analysis (AnchorScore, per-class accuracy of each of the six MLLMs) on a difficulty proxy: the mean accuracy of the \emph{other} five MLLMs on the same class. AnchorScore correlates with this leave-one-out consensus proxy ($\rho=0.763$, $n=78$) even slightly more strongly than with any single held-out MLLM (mean $\rho=0.692$), and its partial correlation controlling for the proxy is close to zero (pooled $\hat{\rho}=0.067$, cluster-bootstrap 95\% CI $[-0.036,0.789]$; class-level $\hat{\rho}=0.219$, $n=13$, n.s.). Thus AnchorScore's predictive content is consistent with a predominantly \emph{shared-difficulty} account (which classes are hard for the ensemble as a whole), though the wide confidence interval prevents quantifying how dominant the shared component is. This matches the marginal-difficulty reading of the Mantel null (\S\ref{sec:mantel}), and the practical implication is that AnchorScore serves as a low-compute, label-seeded cold-start \emph{proxy} of the shared difficulty factor (no MLLM inference; only a small labeled calibration set of about 20 images per class), not a CLIP-specific error source detector.

\textbf{Prompt-randomization control.} The consensus control constrains \emph{what} the association tracks (shared ensemble difficulty vs.\ a CLIP-specific source), but not \emph{where in CLIP} the signal lives. A pure image-side shared-difficulty account (H2) predicts the association is carried by CLIP's image representations and should survive perturbation of the text side; the data-overlap (H1) and semantic-salience (H4) accounts predict it depends on the class--prompt alignment in the text embeddings. We test this by breaking only the class--prompt alignment, with images, backbone identity (ViT-L/14, LAION-2B), prompt templates, the per-dataset protocol, and MLLM accuracies held constant within the inference environment: (i)~\emph{shuffled semantics}, where each dataset's semantic class descriptions are reassigned to class slots by a random derangement (no class keeps its own description; 10 independent seeds), and (ii)~\emph{nonsense}, where non-semantic placeholder prompts replace the class descriptions. Under the canonical condition the recomputed association remains significant in this environment ($\rho=0.648$, $p=0.017$, $n=13$, 95\% CI $[0.152,0.884]$; a 20{,}000-permutation null gives two-sided $p=0.017$). Under shuffled semantics no seed reaches significance (mean $\rho=0.181$ across 10 seeds, range $[-0.105,0.492]$), and under nonsense prompts it vanishes ($\rho=-0.042$, $p=0.89$). Because the image embeddings are identical across the three conditions, the collapse of the association under text-side perturbation argues against a pure image-side shared-difficulty account (H2) and is consistent with the alignment-dependent accounts (H1, H4). Absolute levels in this environment differ modestly from the original run (different CLIP build; \S\ref{sec:limitations}), so only the within-environment contrast is used, which is independent of absolute levels. The control replicates in a second domain: on Stanford40 Actions (40 classes), re-run in the same environment as the canonical AnchorScore computation (self-check: per-class accuracies identical to the cached canonical run, max deviation 0.00~pp), the canonical association reproduces ($\rho=0.817$, 95\% CI $[0.678,0.899]$, $p<0.001$, $n=40$), while under shuffled semantics no seed of 10 shows a significant positive association (mean $\rho=-0.203$, range $[-0.407,0.015]$; the single marginally positive seed has $\rho=+0.015$, $p=0.93$) and under nonsense prompts the association vanishes ($\rho=-0.398$; the between-class SD of AnchorScore falls from 14.0 to 4.4~pp). The class--prompt alignment dependence is therefore not specific to the classroom domain.

\textbf{Four competing hypotheses.} \textbf{Shared pretraining data} (H1) is partially ruled out by the word-frequency partial correlation ($\rho=0.594$ controlling for log word frequency) and the attenuation of the effect on medical data ($\rho=0.214$, $p=0.32$, $n=24$; $\rho=0.046$, $p=0.83$ with the domain-specialized BiomedCLIP backbone), though the latter equally fits H1: neither CLIP (LAION-2B) nor the MLLMs were trained on medical images, so the absence of shared training data would produce a null correlation under either hypothesis. \textbf{Shared visual competence} (H2) is argued against in its image-side form by the prompt-randomization control above (replicated on Stanford40): breaking only the class--prompt alignment destroys the association while the image embeddings are unchanged, so a difficulty signal carried by the image representation alone is insufficient; the vision-only nulls (SigLIP $\rho=0.201$, DINOv2 $\rho=0.050$, ResNet-50 $\rho=0.081$) are consistent with this reading. The two controls are complementary: the consensus control shows what the association \emph{tracks} (ensemble-shared difficulty), and the randomization control indicates that the association depends on class--prompt alignment rather than being determined by image-side representations alone. Together they indicate that CLIP provides a better difficulty proxy than the alternatives, in a manner consistent with its zero-shot score coupling image content to class semantics rather than its image encoder being special.

\textbf{Shared task structure} (H3) is argued against by the supervised ResNet-50 near-zero correlation. \textbf{Semantic salience} (H4), compositional class names being hard for both models, is supported by the failure analysis of \S\ref{sec:main_result}---the largest AnchorScore--MLLM divergences occur on composite action classes---but not systematically tested. No single hypothesis is fully supported or ruled out, but the balance of evidence now favors a shared-difficulty account expressed through class--prompt alignment (H4, with an unknown contribution from H1) over a pure image-side account (H2): the word-frequency partial correlation and the attenuation of the effect on medical data provide the strongest specific evidence against a pure shared-data account, and the prompt-randomization control strongly argues against a purely image-side carrier. Disentangling the remaining candidates---pretraining data overlap (H1) versus semantic salience (H4)---requires controlled experiments where training data and model objective are independently varied.

\textbf{Testing a stronger structural prediction.} A stronger version of H2 would predict that CLIP and MLLMs share not only per-class difficulty but also \emph{which} classes they confuse with \emph{which} others, making CLIP's confusion matrix a powerful MLLM error diagnostic. To test this transitivity hypothesis, symmetrized confusion matrices were built for CLIP and each MLLM on TeacherBehavior (8 classes, 28 class pairs). For each model, $A_{ij}=\tfrac{1}{2}(P(\text{pred}{=}j\!\mid\!\text{true}{=}i)+P(\text{pred}{=}i\!\mid\!\text{true}{=}j))$, and a Mantel test (10{,}000 simultaneous row/column permutations, Spearman) was run between the upper-triangle vectors of CLIP's matrix and each MLLM's matrix.

\label{sec:mantel}
The hypothesis is \emph{not supported}. Per-model Spearman $\rho$ ranged $0.193$--$0.487$ (mean $0.279$), with one uncorrected hit in six tests ($\approx 0.3$ expected under the global null) and none surviving Benjamini--Hochberg correction. A split-half reliability check on CLIP against itself ($\rho{=}0.978$, $p{=}10^{-4}$) confirms the test has power to detect structural agreement when it genuinely exists. This null concerns the \emph{error structure} (which specific class pairs are confused), a different quantity from the shared per-class difficulty that AnchorScore measures: vision encoders may converge on which classes are globally harder while still diverging in the particular pairwise confusions they make, and this distinction is consistent with the finding in \S\ref{sec:cross_domain} that the per-class signal survives across very different encoders even though confusion-level transfer is absent.

One caveat applies to the MLLM side of this analysis: the confusion matrices derive from the same full-frame image-level evaluation pass that produced this collapse, and that pass yields per-class accuracies that diverge substantially from the canonical bbox-cropped evaluation used for the headline correlations (e.g., for Qwen3.5-27B, ``guide'' rises from 52.3\% to 90.4\% and ``blackboard-writing'' falls from 98.2\% to 0\%). The Mantel null therefore characterizes confusion structure under full-frame input; its transfer to the canonical evaluation condition is untested. This null is read as evidence that the AnchorScore--MLLM association is a \emph{marginal-difficulty} effect (the two families agree on which classes are hard) rather than a \emph{shared-error-structure} effect. This interpretation matches the model-specific prompt-disambiguation gains (\S\ref{sec:prompt_opt}), where only the marginal signal (which classes to clarify) transfers, not the pairwise structure.

\subsection{Operating Range}

This subsection first characterizes the AnchorScore range and domains in which the diagnostic carries signal, then addresses the practical considerations for its deployment.

The correlation is strongest for mid-range AnchorScore values (20--60\%; consistent with the tier analysis below). For very low AnchorScore classes (e.g., ``answer'' at 5.6\%, ``blackboard'' at 3.2\%), the relationship is noisier because CLIP's semantic misalignment dominates: MLLMs with strong language understanding can sometimes overcome that misalignment (e.g., ``blackboard'' where MLLM accuracy averaged 44.6\% despite CLIP's 3.2\%). The boundary conditions of this operating range are quantified in the calibration analysis below.

The per-class divergence patterns provide further evidence for the distinction between CLIP's vision-language bridging and MLLMs' more flexible instruction-following. CLIP's nearest-neighbor predictions for blackboard-writing returned ``blackboard'' as the top category, suggesting that the text encoder maps the verb-noun composition to the static object. MLLMs nevertheless achieve high accuracy on this class, suggesting that MLLM performance can sometimes overcome the semantic mismatch reflected in CLIP's zero-shot score. CLIP's systematic underestimation of MLLM accuracy is a uniform bias rather than a class-dependent one (\S\ref{sec:main_result}). Consequently, practitioners should note that AnchorScore tended to underestimate MLLM accuracy for the composite action classes in SCB5, while remaining a useful relative ranking signal for simple noun classes.

The cross-domain results reveal an additional boundary: AnchorScore's diagnostic value is domain-dependent. The per-class correlation is strongest in domains where CLIP has moderate visual-linguistic competence (e.g., classroom scenes at $\rho=0.769$), not where it performs very well (satellite) or very poorly (medical domains, where the pooled within-medical signal is non-significant: $\rho=0.214$, $p=0.32$, $n=24$). The between-domain ordering is also partially inconsistent: PathMNIST (AnchorScore 15.5\%) shows lower MLLM accuracy than BloodMNIST (AnchorScore 5.9\%), the opposite of what AnchorScore alone would predict. The pooled cross-domain correlation ($\rho=0.462$, $p=0.006$, $n=34$; \Cref{tab:cross_mllm}) arises from between-domain level differences when pooling across datasets, not from strong within-domain ranking in any single dataset (per-dataset within-domain correlations are all non-significant). This attenuation is consistent with domain dissimilarity: EuroSAT and MedMNIST differ from classroom behavior not only in visual appearance but in underlying task structure (land-cover and cellular classification, not human action recognition). These differences likely reflect varying overlap with CLIP's web-scale pretraining distribution, though this was not tested directly.

Two empirical patterns emerge from the per-class data. First, the class-level correlation is strongest when AnchorScore values span a wide range within a single dataset (e.g., SCB5, where 13 classes span 3--90\%), giving the ranker sufficient leverage. Second, when AnchorScore values cluster at the extremes (either uniformly low, as in the medical datasets all $<16\%$, or uniformly high, as in Stanford40 where 37 of 40 classes exceed 60\%), within-dataset ranking has less room to discriminate, and the between-dataset signal attenuates accordingly. Across datasets, classes with AnchorScore $<$10\% ($n=9$, spanning SCB5 low-accuracy and BloodMNIST) show high inter-class variance in MLLM accuracy; classes in 20--60\% ($n=10$, SCB5 mid-range plus EuroSAT) drive the bulk of the class-level $\rho=0.769$; and classes $>$60\% ($n=40$, SCB5 high-accuracy plus Stanford40) operate in a ceiling region where both models perform well. These tiers (using different cutoffs than the three-tier heuristic of the calibration analysis below) are illustrative observations about where the diagnostic has empirical leverage, not validated thresholds; determining the AnchorScore range where the diagnostic adds value in a new domain requires inspecting the per-class spread in that domain directly.

The remaining considerations concern the precision and stability of the diagnostic in practice.

\textbf{Calibration analysis.} The overall Expected Calibration Error (ECE, 5 bins) is 0.152 (see \Cref{fig:calibration} in Appendix~\ref{app:materials}), confirming that AnchorScore is not a calibrated predictor of MLLM accuracy in absolute terms. A three-tier heuristic ($<10\%$ low, $10$--$40\%$ mid, $>40\%$ high) captures the central tendency but not the spread: pooled across datasets, classes with AnchorScore $\leq 10\%$ have mean MLLM accuracy of 18.6\% ($n=23$) versus 64.7\% ($n=13$) for classes $>40\%$; on the 13 SCB5 classes alone, low-tier classes ($n=3$) average $26.7\%$ (range $12.4$--$44.6\%$), mid-tier ($n=3$) $55.6\%$ ($27.0$--$97.5\%$), and high-tier ($n=7$) $81.1\%$ ($67.8$--$97.8\%$), with all three low-tier classes below $50\%$---reliable candidates for mandatory human review. The per-class spread remains large in both directions (e.g., ``blackboard-writing'' with $\Delta=+71.4$~pp and ``PermanentCrop'' with $\Delta=-77.3$~pp), reflecting cases where CLIP's semantic alignment diverges substantially from MLLM visual understanding. In practice, AnchorScore supports \emph{ranking} (identifying which classes are relatively harder), not \emph{calibration} (estimating the exact accuracy).

\textbf{Validation set size requirements.} A practical question is how many labeled validation images per class are needed for AnchorScore to stabilize. The canonical per-image CLIP prediction cache is bootstrap-resampled at $N\in\{5,10,20,50,100,\text{all}\}$ images per class ($B=200$ repetitions) and the class-level $\rho$ is computed at each $N$. The correlation stabilizes quickly: at $N=5$, $\rho=0.669\pm0.124$; at $N=20$, $\rho=0.753\pm0.062$; and at $N=50$, $\rho=0.771\pm0.036$ (at $N=\text{all}$ it reproduces the headline $\rho=0.769$ exactly). Beyond $N=50$, the confidence interval narrows only marginally. This indicates that approximately 20 labeled images per class sufficed for a stable AnchorScore estimate on SCB5---far fewer than typical validation sets.

\textbf{Prompt robustness.} As detailed in the prompt construction guideline (\S\ref{sec:method}), domain context is essential for the signal and specific phrasing is flexible; the original 3-template prompt matches the headline result exactly, and all variants are tabulated in Appendix~\ref{app:materials} (\Cref{tab:prompt_robustness}). Note that this makes AnchorScore training-free but not prompt-effort-free: deploying on a new domain requires constructing a domain-context prompt, and some prompt-engineering trial-and-error is part of that deployment cost.

\textbf{Deployment summary.} Combining the above, the pre-deployment procedure is: (i)~verify that the target domain has moderate CLIP competence and a wide within-dataset AnchorScore spread---floor regimes (medical) and ceiling regimes (Stanford40) offer little ranking leverage; (ii)~label $\geq$20 images per class and construct a domain-context prompt, checking the per-class spread on the labeled set; (iii)~apply the resulting AnchorScore ranking to route images with a budget-determined threshold $\tau$, flag the bottom $k/3$ classes for human review, or steer prompt disambiguation---always treating AnchorScore as a ranking signal, not a calibrated accuracy estimate.

\subsection{Threats to Validity}
\label{sec:limitations}

The following scope conditions apply to the interpretation of the results.

(1) \textbf{Statistical power.} The primary SCB5 analysis ($n=13$) includes two sub-datasets with only 2--3 classes where Spearman $\rho$ has a structural ceiling. The TeacherBehavior subset ($k=8$) alone yields $\rho=0.595$ ($p=0.12$)---directionally consistent but underpowered. Significance at the class level is achieved only on the pooled SCB5 set and in the cross-domain per-class analysis; the classroom replication (SCB-LLM, $\rho=0.506$, $p=0.135$) is individually non-significant. The synthesis prediction interval $[0.126, 0.909]$ acknowledges between-study uncertainty. Among the alternative metrics and subsets evaluated, AnchorScore is the only predictor surviving Benjamini--Hochberg correction; CLIP inverse entropy reaches nominal significance ($\rho=0.604$, $p=0.029$) but does not survive correction, and all remaining alternatives are non-significant: alternative vision models (SigLIP $\rho=0.201$, DINOv2 $\rho\approx0.05$, ResNet-50 $\rho\approx0.08$), MLLM self-uncertainty ($\rho=-0.07$), CLIP confidence/margin ($\rho=0.434$, $p=0.14$), dispersion ($\rho=0.203$, $p=0.51$), and subset or backbone variants (OpenAI L/14 $\rho=0.473$, B/32 $\rho=0.242$, TeacherBehavior alone $\rho=0.595$, SCB-LLM $\rho=0.506$). AnchorScore was the only predictor surviving FDR correction within the evaluated alternative-predictor family. Beyond the confirmatory families (the eight predictors, LOCO, and Mantel, all BH-corrected), the remaining comparisons across backbones, domains, prompt variants, and routing thresholds are exploratory, and their individual $p$-values should be read as descriptive rather than inferential.

(2) \textbf{Domain dependency.} The strongest signal is on classroom behavior data. The per-class cross-domain correlation ($\rho=0.462$, $p=0.006$, $n=34$) is directionally consistent with but weaker than the primary SCB5 result, and the cross-domain between-domain ordering is partially inconsistent; the cross-domain MLLMs are 7B models, so the attenuation is partly confounded with annotator strength, though a 27B scale-check on EuroSAT leaves the between-domain ordering unchanged (\S\ref{sec:cross_domain}). The prompt-disambiguation analysis (\S\ref{sec:prompt_opt}) is validated only on SCB5; the routing and ranking analyses extend to Stanford40. Transferability of all three to other domains remains to be tested. The routing estimate additionally assumes that MLLM accuracy on routed images equals the canonical per-class rate; realized runs under the canonical protocol validate this estimate within 1.8~pp, and a full-frame deployment-condition run quantifies the protocol sensitivity (\S\ref{sec:hybrid}).

(3) \textbf{MLLM and backbone scope.} The MLLMs span Qwen, Gemma, and LLaVA (7B--35B), all open-source decoder-only VLMs. Whether AnchorScore predicts accuracy for proprietary models (GPT-4V, Gemini Pro) or encoder-decoder architectures is untested. The class-level association reaches significance only for LAION-pretrained ViT-L/14; OpenAI variants are directionally consistent but individually non-significant ($\rho=0.24$--$0.47$, $p>0.1$). The six evaluated MLLMs, while nominally distinct, exhibit substantial pairwise agreement in per-class difficulty rankings (mean pairwise $\rho=0.864$, range $[0.764, 0.956]$ across all 15 model pairs, $n=13$). This does not affect the class-level correlation analysis, which uses the cross-model mean, but it weakens any robustness argument that relies on treating the six MLLMs as independent replicates (most notably the leave-one-MLLM-out analysis in \S\ref{sec:main_result}, where holding out a highly correlated model provides less independent confirmation than the nominal model count suggests). Future work should incorporate MLLMs that differ not only in scale or family but in evaluation paradigm (e.g., open-ended generative scoring vs.\ multiple-choice selection) to more rigorously test whether AnchorScore's predictive value generalizes beyond a shared assessment format. Logit-based uncertainty baselines~\cite{kadavath2022language} remain unevaluated; verbalized uncertainty is evaluated in \S\ref{sec:extended_baselines} and shows no class-level signal.

(4) \textbf{Inherited biases.} CLIP encodes markedness-driven social asymmetries~\cite{wolfe2022clip}, and AnchorScore may inherit these. Two factors partially mitigate the practical impact in this setting. First, classroom behavior classification is action-based (writing, reading, hand-raising, standing) rather than identity-based; each class contains images of students across demographic groups, so per-group accuracy differences are partially diluted at the class-aggregate level. Second, the downstream applications (\S\ref{sec:applications}) use AnchorScore to rank classes, not to classify individual images, which further reduces per-instance bias amplification. However, practitioners deploying AnchorScore on datasets with demographic relevance (particularly those involving person detection) should audit CLIP's per-group accuracy for systematic disparities before using the ranking to allocate review resources.

(5) \textbf{Task scope.} Results are restricted to single-label image classification. Extension to detection, segmentation, or multi-label tasks requires additional validation. AnchorScore also requires $\geq$20 labeled validation images per class; cold-start scenarios with fewer labels will have noisier AnchorScore estimates, as shown in the validation-size ablation (\Cref{sec:discussion}).

(6) \textbf{Mechanism attribution and baseline feasibility.} The consensus control (\S\ref{sec:consensus_control}) shows that the pooled partial association of AnchorScore given the other-MLLM consensus proxy is close to zero, but the cluster-bootstrap 95\% CI is wide ($[-0.036,0.789]$, $B=10{,}000$) and the class-level partial ($\hat{\rho}=0.219$, $n=13$) is non-significant; the data therefore support the \emph{shared-difficulty proxy} framing but cannot rule out a substantial CLIP-specific component (the CI's upper bound of 0.789 approaches the raw pooled correlation). The null results of the alternative cheap predictors are additionally constrained by their accuracy floors: among the 13 SCB5 classes, BLIP-2 and ResNet-50 compress 69\% of classes below 15\% accuracy, and BLIP-2 assigns 15\% of classes at or above 85\%, leaving little rank-discriminating variance where the alternative-predictor nulls are claimed. ResNet-50 in particular (mean 12.5\%, max 62.1\%) operates almost entirely in the floor. These floor effects partially confound the baseline nulls: a weak predictor with no usable range cannot display a correlation. The baseline comparisons are therefore presented as \emph{necessary} conditions---AnchorScore is the cheapest predictor that clears the floor---rather than as proof of an exclusive CLIP mechanism. The prompt-randomization control (\S\ref{sec:consensus_control}) carries three scope conditions of its own: (i) it was run under a different CLIP build than the original canonical run (self-check against the cached per-class accuracies: mean $9.0$~pp, max $32.3$~pp difference), so only the within-environment contrast (canonical vs.\ shuffled vs.\ nonsense) is claimed, which is independent of absolute levels; (ii) for HandriseReadWrite (3 classes) and BowTurnHead (2 classes) only 2 and 1 distinct derangements exist, so per-seed diversity of the shuffled condition is limited on those sub-datasets (the shuffled-condition collapse is driven mainly by TeacherBehavior); and (iii) the nonsense condition does not fully collapse per-class accuracy to chance (several classes retain high accuracy under non-semantic prompts), so the floor check is partial---the result used is the collapse of the \emph{correlation with MLLM accuracy}, not of absolute accuracy. The control's domain scope is now two-fold: the Stanford40 replication (same environment, max self-check deviation 0.00~pp) shows the same class--prompt dependency in activity-recognition data, though further domains remain untested.

\section{Conclusions}
\label{sec:conclusion}

This paper presented AnchorScore, a training-free diagnostic that uses a frozen CLIP model's zero-shot accuracy to flag which classes MLLMs are least likely to annotate reliably, at roughly 1/270th of the estimated per-image inference compute of direct MLLM evaluation. On classroom behavior data, AnchorScore correlated with per-class MLLM accuracy at $\rho=0.769$ ($p=0.002$). The signal was present only for CLIP among the evaluated cheap predictors: SigLIP, DINOv2, ResNet-50, and MLLM self-uncertainty all showed no significant class-level correlation at $n=13$ (several near accuracy floors), and a small-MLLM pilot at a matched label budget carries no class-level signal. A cross-model consensus control indicates the signal is consistent with a \emph{shared} class-difficulty factor: conditioning on the other five MLLMs' consensus leaves a partial correlation near zero ($\hat{\rho}=0.067$, n.s.; 95\% CI $[-0.036, 0.789]$ cannot exclude a substantial CLIP-specific component), so AnchorScore's value lies in being a low-compute, label-seeded cold-start proxy of shared difficulty rather than a CLIP-exclusive mechanism. An independent replication on Stanford40 Actions (40 classes, 6 MLLMs) yielded an almost identical effect ($\rho=0.817$, $p<0.001$), with a three-dataset activity-recognition synthesis producing a pooled estimate of $\rho=0.781$ ($I^2=3.0\%$).
 
AnchorScore is not a universal predictor. The effect was strong and consistent on activity-recognition tasks ($\rho=0.769$ to $0.817$, $I^2=3.0\%$) but attenuated on medical and satellite domains (broader synthesis $\rho=0.677$, $I^2=62.3\%$). Three downstream analyses showed that the diagnostic signal can translate into practical gains: deployable predicted-class routing improved accuracy by up to 23~pp over CLIP-only at about 44\% MLLM cost savings under the canonical MLLM evaluation, with realized runs of $+$21.7~pp (canonical protocol) and $+$17.5~pp (full-frame deployment condition; \S\ref{sec:hybrid}), prompt disambiguation yielded positive but model-specific gains on hard classes (mean $+25.0$~pp for direct visual-distinction descriptions; confuser controls localize these gains to the two lowest-AnchorScore classes and to the CLIP-derived pair specifically, read as exploratory), and AnchorScore ranking identified review-worthy classes on both TeacherBehavior and Stanford40 (AUC 0.84--0.94). A prompt-randomization control---breaking only the class--prompt alignment with images, backbone identity, templates, and protocol held constant within the environment---further localizes the signal: the association collapses under shuffled semantics (10-seed mean $\rho=0.181$) and nonsense prompts ($\rho=-0.042$), both non-significant against the canonical $\rho=0.648$ ($p=0.017$) computed in the same environment, arguing against a pure image-side difficulty account and consistent with the alignment-dependent accounts (data overlap, semantic salience); the control replicates on Stanford40, where the canonical association ($\rho=0.817$, $p<0.001$) likewise collapses under shuffled semantics (mean $\rho=-0.203$; no positive seed of 10) and nonsense prompts ($\rho=-0.398$). Future work should test transfer to further domains, proprietary MLLMs, and open-ended evaluation paradigms, and should separate the remaining candidates---pretraining data overlap versus semantic salience---by independently varying training data and model objective. With a small labeled calibration set (empirically, about 20 images per class in SCB5) and 3 minutes of CLIP inference, practitioners can obtain a per-class difficulty map that enables smarter resource allocation in MLLM annotation pipelines.


\appendix
\setcounter{figure}{0}
\renewcommand{\thefigure}{A\arabic{figure}}
\setcounter{table}{0}
\renewcommand{\thetable}{A\arabic{table}}
\clearpage
\section{Per-Class Evidence and Diagnostics}
\label{app:materials}

\begin{figure}[t]
\centering
\includegraphics[width=\columnwidth]{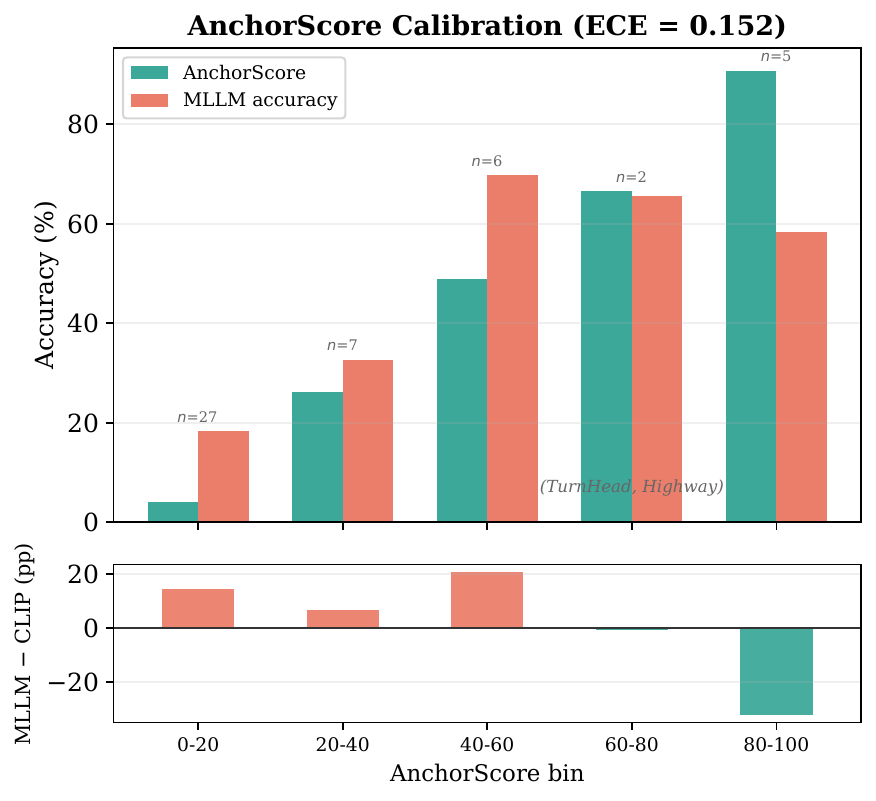}
\caption{Calibration plot: AnchorScore vs.\ mean MLLM accuracy across 47 classes (5 bins); perfect calibration would equalize bar heights per bin. The lower panel quantifies the gap (MLLM $-$ CLIP in pp); ECE $=$ 0.152, consistent with AnchorScore supporting ranking rather than absolute accuracy estimation.}
\label{fig:calibration}
\end{figure}

This appendix provides the full per-class evidence underlying the main text: the 13-class SCB5 breakdown (Table~\ref{tab:per_class_tb}), the 40-class Stanford40 breakdown (Table~\ref{tab:stanford40_per_class}), the 34-class cross-domain breakdown (Table~\ref{tab:cross_domain_per_class}), the prompt-variant robustness analysis (Table~\ref{tab:prompt_robustness}), and the calibration diagnostics (Fig.~\ref{fig:calibration}). Throughout this appendix, $\Delta$ denotes mean MLLM accuracy minus AnchorScore (percentage points); positive values mean the MLLM outperforms CLIP on that class.

\textbf{SCB5 per-class accuracy.} \Cref{tab:per_class_tb} reports AnchorScore and per-class MLLM accuracy (6-MLLM mean $\pm$ SD) for all 13 SCB5 classes across the three sub-datasets, sorted by ascending AnchorScore. The largest AnchorScore--MLLM divergences occur on composite action classes (blackboard-writing +71.4~pp, blackboard +41.4~pp). The companion paper reports per-class accuracy for a different model set and filtering strategy (Qwen2-VL-7B and LLaVA-1.5-7B under the Agreement strategy on bbox-cropped inputs)~\cite{ma2026agreement}; the two per-class tables are complementary and not directly comparable.

\begin{table}[t]
\caption{Per-class AnchorScore and average MLLM accuracy across all 13 SCB5 classes (TeacherBehavior, HandriseReadWrite, BowTurnHead; 6 MLLMs), grouped by sub-dataset and sorted by ascending AnchorScore. MLLM std is the sample standard deviation of the six per-model accuracies.}
\label{tab:per_class_tb}
\centering
\footnotesize
\setlength{\tabcolsep}{2.8pt}
\begin{tabular}{lcccc}
\toprule
\textbf{Class} & \textbf{Anchor\%} & \textbf{MLLM avg\%} & \textbf{MLLM std\%} & $\Delta$\,(pp) \\
\midrule
\multicolumn{5}{l}{\textit{TeacherBehavior}} \\
blackboard & 3.2 & 44.6 & 15.6 & $+$41.4 \\
answer & 5.6 & 12.4 & 7.6 & $+$6.8 \\
stand & 9.6 & 23.1 & 6.7 & $+$13.5 \\
blackboard-writing & 26.1 & 97.5 & 1.9 & $+$71.4 \\
guide & 29.3 & 42.5 & 22.6 & $+$13.2 \\
screen & 40.5 & 67.8 & 15.0 & $+$27.3 \\
teacher & 50.5 & 68.8 & 8.8 & $+$18.3 \\
On-stage interaction & 54.2 & 80.2 & 9.0 & $+$26.0 \\
\midrule
\multicolumn{5}{l}{\textit{HandriseReadWrite}} \\
read & 23.0 & 27.0 & 5.7 & $+$4.0 \\
hand-raising & 80.0 & 89.2 & 2.6 & $+$9.2 \\
write & 89.8 & 97.8 & 0.4 & $+$8.0 \\
\midrule
\multicolumn{5}{l}{\textit{BowTurnHead}} \\
BowHead & 49.1 & 80.5 & 17.9 & $+$31.4 \\
TurnHead & 64.1 & 83.7 & 9.4 & $+$19.6 \\
\bottomrule
\end{tabular}
\end{table}

\textbf{Stanford40 per-class accuracy.} \Cref{tab:stanford40_per_class} reports per-class AnchorScore and mean MLLM accuracy (6 MLLMs, 50 images per class each) for all 40 Stanford40 classes, sorted by ascending AnchorScore. The table substantiates the ceiling-regime observation in \S\ref{sec:stanford40}: 37 of 40 classes exceed 60\% AnchorScore, and the correlation is driven primarily by the lowest-AnchorScore classes (cutting vegetables, texting message, writing on a book).

\begin{table}[t]
\caption{Per-class AnchorScore and average MLLM accuracy on Stanford40 (40 classes, 6 MLLMs), sorted by ascending AnchorScore. Left panel: classes 1--20; right panel: classes 21--40. MLLM std omitted for space.}
\label{tab:stanford40_per_class}
\centering
\footnotesize
\setlength{\tabcolsep}{3pt}
\begin{tabular}{lccc@{\hspace{1.8em}}lccc}
\toprule
\textbf{Class} & \textbf{Anchor\%} & \textbf{MLLM\%} & $\Delta$ & \textbf{Class} & \textbf{Anchor\%} & \textbf{MLLM\%} & $\Delta$ \\
\midrule
cutting vegetables & 47.6 & 81.3 & $+$33.7 & reading & 97.6 & 96.0 & $-$1.6 \\
texting message & 52.9 & 90.3 & $+$37.5 & cooking & 97.6 & 99.3 & $+$1.8 \\
writing on a book & 58.9 & 89.7 & $+$30.7 & watching TV & 97.8 & 98.7 & $+$0.9 \\
pouring liquid & 61.0 & 96.3 & $+$35.3 & holding an umbrella & 98.0 & 99.7 & $+$1.7 \\
applauding & 69.4 & 91.3 & $+$22.0 & throwing frisbee & 98.0 & 98.0 & $+$0.0 \\
waving hands & 78.1 & 99.3 & $+$21.2 & riding a bike & 98.3 & 100.0 & $+$1.7 \\
brushing teeth & 79.0 & 91.7 & $+$12.7 & writing on a board & 98.4 & 99.7 & $+$1.3 \\
taking photos & 85.8 & 95.3 & $+$9.5 & feeding a horse & 98.6 & 100.0 & $+$1.4 \\
washing dishes & 86.3 & 95.3 & $+$9.1 & playing violin & 98.9 & 100.0 & $+$1.2 \\
looking through a telescope & 89.7 & 95.3 & $+$5.7 & climbing & 99.0 & 100.0 & $+$1.0 \\
smoking & 93.0 & 96.0 & $+$3.1 & fixing a bike & 99.1 & 100.0 & $+$0.9 \\
drinking & 94.9 & 98.0 & $+$3.1 & fishing & 99.3 & 100.0 & $+$0.7 \\
using a computer & 95.2 & 98.7 & $+$3.4 & riding a horse & 99.3 & 100.0 & $+$0.7 \\
jumping & 95.6 & 96.3 & $+$0.7 & walking the dog & 99.3 & 100.0 & $+$0.7 \\
phoning & 95.8 & 97.0 & $+$1.3 & looking through a microscope & 99.5 & 98.0 & $-$1.5 \\
cleaning the floor & 96.7 & 99.7 & $+$3.0 & cutting trees & 99.5 & 97.3 & $-$2.2 \\
running & 96.8 & 98.3 & $+$1.5 & shooting an arrow & 99.5 & 100.0 & $+$0.5 \\
blowing bubbles & 96.9 & 97.3 & $+$0.4 & playing guitar & 99.7 & 100.0 & $+$0.4 \\
gardening & 97.0 & 99.0 & $+$2.0 & fixing a car & 100.0 & 100.0 & $+$0.0 \\
pushing a cart & 97.5 & 99.0 & $+$1.6 & rowing a boat & 100.0 & 100.0 & $+$0.0 \\
\bottomrule
\end{tabular}
\end{table}

\textbf{Cross-domain per-class accuracy.} \Cref{tab:cross_domain_per_class} reports the 34 cross-domain classes underlying the pooled analysis in \S\ref{sec:pooled_meta} (13 SCB5 classes $+$ 34 cross-domain classes $=$ 47), grouped by dataset and sorted by ascending AnchorScore, with each dataset's within-dataset Spearman $\rho$. For TissueMNIST, the merged AnchorScore class ``Collecting Duct, Connecting Tubule'' is compared against the MLLM's ``Collecting Duct'' accuracy only, because the MLLM evaluation separated the two sub-classes, and ``Thick Ascending Limb'' (no MLLM counterpart) is excluded. The table substantiates the domain-dependence claim: no within-dataset correlation reaches significance (TissueMNIST's $\rho=0.741$ is strong but $n=7$), and the pooled signal arises from between-domain level differences.

\begin{table}[t]
\caption{Per-class AnchorScore and mean MLLM accuracy (4 MLLMs) across the four cross-domain datasets, sorted by ascending AnchorScore within each dataset. MLLM std is the sample standard deviation of the four per-model accuracies.}
\label{tab:cross_domain_per_class}
\centering
\footnotesize
\setlength{\tabcolsep}{2.8pt}
\begin{tabular}{lcccc}
\toprule
\textbf{Class} & \textbf{Anchor\%} & \textbf{MLLM avg\%} & \textbf{MLLM std\%} & $\Delta$ \\
\midrule
\multicolumn{5}{l}{\textit{EuroSAT (10 classes; within-dataset $\rho=0.286$)}} \\
HerbaceousVegetation & 0.3 & 1.0 & 1.2 & $+$0.7 \\
AnnualCrop & 0.5 & 36.5 & 41.5 & $+$36.0 \\
Forest & 2.5 & 56.0 & 24.7 & $+$53.5 \\
Pasture & 13.7 & 36.5 & 41.8 & $+$22.9 \\
SeaLake & 23.7 & 41.0 & 36.2 & $+$17.3 \\
River & 41.8 & 65.5 & 15.6 & $+$23.7 \\
Industrial & 57.7 & 55.5 & 8.9 & $-$2.2 \\
Highway & 69.0 & 47.5 & 37.0 & $-$21.5 \\
PermanentCrop & 90.8 & 13.5 & 27.0 & $-$77.3 \\
Residential & 96.0 & 49.5 & 16.1 & $-$46.5 \\
\midrule
\multicolumn{5}{l}{\textit{PathMNIST (9 classes; within-dataset $\rho=0.162$)}} \\
debris & 0.0 & 0.5 & 1.0 & $+$0.5 \\
lymphocytes & 0.0 & 2.5 & 3.0 & $+$2.5 \\
mucus & 0.0 & 3.0 & 6.0 & $+$3.0 \\
cancer-associated stroma & 0.0 & 35.0 & 42.9 & $+$35.0 \\
normal colon mucosa & 2.0 & 17.5 & 18.1 & $+$15.5 \\
adipose & 4.8 & 25.5 & 45.8 & $+$20.7 \\
colorectal adenocarcinoma epithelium & 10.5 & 0.0 & 0.0 & $-$10.5 \\
smooth muscle & 13.3 & 2.5 & 5.0 & $-$10.8 \\
background & 97.3 & 42.0 & 48.5 & $-$55.3 \\
\midrule
\multicolumn{5}{l}{\textit{BloodMNIST (8 classes; within-dataset $\rho=0.019$)}} \\
erythroblast & 0.0 & 0.0 & 0.0 & $+$0.0 \\
neutrophil & 0.0 & 24.0 & 28.0 & $+$24.0 \\
platelet & 0.2 & 24.0 & 23.4 & $+$23.8 \\
immature granulocytes & 4.7 & 0.0 & 0.0 & $-$4.7 \\
basophil & 6.2 & 24.5 & 38.8 & $+$18.4 \\
eosinophil & 7.7 & 48.0 & 43.5 & $+$40.3 \\
lymphocyte & 20.6 & 2.5 & 2.5 & $-$18.1 \\
monocyte & 21.1 & 0.0 & 0.0 & $-$21.1 \\
\midrule
\multicolumn{5}{l}{\textit{TissueMNIST (7 classes; within-dataset $\rho=0.741$)}} \\
Glomerular endothelial cells & 0.0 & 0.0 & 0.0 & $+$0.0 \\
Interstitial endothelial cells & 0.0 & 0.0 & 0.0 & $+$0.0 \\
Proximal Tubule Segments & 1.8 & 2.0 & 2.3 & $+$0.2 \\
Leukocytes & 5.5 & 0.0 & 0.0 & $-$5.5 \\
Podocytes & 6.4 & 47.0 & 54.3 & $+$40.6 \\
Distal Convoluted Tubule & 12.6 & 28.0 & 48.3 & $+$15.4 \\
Collecting Duct, Connecting Tubule & 39.2 & 17.5 & 33.7 & $-$21.7 \\
\bottomrule
\end{tabular}
\end{table}

\textbf{Prompt robustness.} \Cref{tab:prompt_robustness} summarizes the class-level Spearman $\rho$ obtained with six prompt variants. Domain-context prompts (rows 1--4) all yield significant correlations; removing domain context (rows 5--6) collapses the signal to non-significance ($\Delta\rho=0.620$).

\begin{table}[t]
\caption{Prompt robustness: class-level Spearman $\rho$ between AnchorScore and MLLM accuracy under six prompt variants on SCB5 ($n=13$, 6-MLLM subset).}
\label{tab:prompt_robustness}
\centering
\footnotesize
\begin{tabular}{lcccl}
\toprule
\textbf{Prompt variant} & $T$ & \textbf{$\rho$} & \textbf{$p$} & \textbf{Domain context?} \\
\midrule
\textbf{3-template average (original)} & 3 & \textbf{0.769} & \textbf{0.002} & Yes \\
``a photo of \{cls\} in a classroom'' & 1 & 0.736 & 0.004 & Yes \\
Single: ``a person \{cls\} in classroom'' & 1 & 0.670 & 0.012 & Yes \\
Alternative 3-template phrasing & 3 & 0.599 & 0.031 & Yes \\
\midrule
``a photo of \{cls\}.'' & 1 & 0.238 & 0.435 & \textbf{No} \\
``a photo of a person \{cls\}.'' & 1 & 0.149 & 0.627 & \textbf{No} \\
\bottomrule
\end{tabular}
\\[2pt]
\footnotesize{$T$ = number of prompt templates averaged. \textbf{Bold} = best-performing variant. All $p$-values from Spearman rank correlation, two-sided, $n=13$.}
\end{table}

\textbf{Calibration.} \Cref{fig:calibration} shows the calibration plot of AnchorScore against mean MLLM accuracy (47 classes, 5 bins, ECE~=~0.152; per-bin class counts from low to high AnchorScore: $n=27$, $7$, $6$, $2$, $5$).

\clearpage


\section*{Data Availability}
The SCB5 datasets used in this study are publicly available on Hugging Face at \url{https://huggingface.co/datasets/wintonYF/SCB-Dataset}. EuroSAT is available at \url{https://github.com/phelber/EuroSAT}. MedMNIST datasets are available at \url{https://medmnist.com}. The experiment code, prompt templates, per-image MLLM predictions, and all result files are available at \url{https://github.com/zhanglizhuo/VisualAnchor}.

\section*{Acknowledgments}
The authors thank the contributors of the SCB dataset, EuroSAT, and MedMNIST for making their data publicly available. During the preparation of this work, the authors used AI-assisted tools for code development and manuscript revision support. After using these tools, the authors reviewed and edited the content as needed and take full responsibility for the content of the publication.

\section*{Declaration of competing interests}
The authors declare no conflicts of interest.


\bibliographystyle{plainnat}
\bibliography{refs}

@inproceedings{liu2023visual,
  author    = {Liu, H and Li, C and Wu, Q and Lee, Y.J},
  title     = {Visual Instruction Tuning},
  booktitle = {Advances in Neural Information Processing Systems (NeurIPS)},
  year      = {2023},
}

@misc{kadavath2022language,
  author    = {Kadavath, S and Conerly, T and Askell, A and Henighan, T and Drain, D and Perez, E and et al},
  title     = {Language Models (Mostly) Know What They Know},
  howpublished = {arXiv:2207.05221},
  year      = {2022},
}

@inproceedings{radford2021clip,
  author    = {Radford, A and Kim, J.W. and Hallacy, C and Ramesh, A and Goh, G and Agarwal, S and Sastry, G and Askell, A and Mishkin, P and Clark, J and Krueger, G and Sutskever, I},
  title     = {Learning Transferable Visual Models From Natural Language Supervision},
  booktitle = {38th International Conference on Machine Learning (ICML)},
  pages     = {8748--8763},
  year      = {2021},
}

@inproceedings{hessel2021clipscore,
  author    = {Hessel, J and Holtzman, A and Forbes, M and Le Bras, R and Choi, Y},
  title     = {CLIPScore: A Reference-free Evaluation Metric for Image Captioning},
  booktitle = {2021 Conference on Empirical Methods in Natural Language Processing (EMNLP)},
  pages     = {7514--7528},
  year      = {2021},
}

@misc{scb_dataset,
  author    = {WintonYF},
  title     = {{SCB-Dataset}: Smart Classroom Behavior Dataset},
  howpublished = {Hugging Face Datasets, \url{https://huggingface.co/datasets/wintonYF/SCB-Dataset}},
  year      = {2025},
  note      = {Dataset.},
}

@inproceedings{schuhmann2022laion5b,
  author    = {Schuhmann, C and Beaumont, R and Vencu, R and Gordon, C and Wightman, R and Cherti, M and Coombes, T and Katta, A and Mullis, C and Wortsman, M and et~al},
  title     = {LAION-5B: An Open Large-Scale Dataset for Training Next Generation Image--Text Models},
  booktitle = {Advances in Neural Information Processing Systems 35 (NeurIPS 2022 Datasets and Benchmarks Track)},
  pages     = {25278--25294},
  year      = {2022},
}

@inproceedings{fang2024data,
  author    = {Fang, A and Jose, A M and Jain, A and Schmidt, L and Toshev, A and Shankar, V},
  title     = {Data Filtering Networks},
  booktitle = {International Conference on Learning Representations (ICLR)},
  year      = {2024},
}

@inproceedings{cherti2023reproducible,
  author    = {Cherti, M and Beaumont, R and Wightman, R and Wortsman, M and Ilharco, G and Gordon, C and Schuhmann, C and Schmidt, L and Jitsev, J},
  title     = {Reproducible Scaling Laws for Contrastive Language--Image Learning},
  booktitle = {IEEE/CVF Conference on Computer Vision and Pattern Recognition (CVPR)},
  pages     = {2818--2829},
  year      = {2023},
}

@article{helber2019eurosat,
  author    = {Helber, P and Bischke, B and Dengel, A and Borth, D},
  title     = {EuroSAT: A Novel Dataset and Deep Learning Benchmark for Land Use and Land Cover Classification},
  journal   = {IEEE J. Sel. Top. Appl. Earth Obs. Remote Sens.},
  volume    = {12},
  number    = {7},
  pages     = {2217--2226},
  year      = {2019},
}

@article{yang2023medmnist,
  author    = {Yang, J and Shi, R and Wei, D and Liu, Z and Zhao, L and Ke, B and Pfister, H and Ni, B},
  title     = {MedMNIST v2: A Large-Scale Lightweight Benchmark for 2D and 3D Biomedical Image Classification},
  journal   = {Sci. Data},
  volume    = {10},
  pages     = {41},
  year      = {2023},
}

@article{oquab2024dinov2,
  author    = {Oquab, M and Darcet, T and Moutakanni, T and Vo, H and Szafraniec, M and Khalidov, V and Fernandez, P and Haziza, D and Massa, F and El-Nouby, A and et~al},
  title     = {DINOv2: Learning Robust Visual Features Without Supervision},
  journal   = {Transactions on Machine Learning Research},
  year      = {2024},
}

@inproceedings{rohrbach2018chair,
  author    = {Rohrbach, A and Hendricks, L.A and Burns, K and Darrell, T and Saenko, K},
  title     = {Object Hallucination in Image Captioning},
  booktitle = {Conference on Empirical Methods in Natural Language Processing (EMNLP)},
  pages     = {4035--4045},
  year      = {2018},
}

@inproceedings{li2023pope,
  author    = {Li, Y and Du, Y and Zhou, K and Wang, J and Zhao, W.X and Wen, J.-R},
  title     = {Evaluating Object Hallucination in Large Vision-Language Models},
  booktitle = {Conference on Empirical Methods in Natural Language Processing (EMNLP)},
  pages     = {292--305},
  year      = {2023},
}

@techreport{settles2009active,
  author    = {Settles, B},
  title     = {Active Learning Literature Survey},
  institution = {University of Wisconsin--Madison, Department of Computer Sciences},
  year      = {2009},
}

@inproceedings{sener2018active,
  author    = {Sener, O and Savarese, S},
  title     = {Active Learning for Convolutional Neural Networks: A Core-Set Approach},
  booktitle = {International Conference on Learning Representations (ICLR)},
  year      = {2018},
}

@misc{burns2023weak,
  author    = {Burns, C and Izmailov, P and Kirchner, J.H and Baker, B and Gao, L and Aschenbrenner, L and Chen, Y and Ecoffet, A and Joglekar, M and Leike, J and Sutskever, I and Wu, J},
  title     = {Weak-to-Strong Generalization: Eliciting Strong Capabilities With Weak Supervision},
  howpublished = {arXiv:2312.09390},
  year      = {2023},
}

@inproceedings{dubois2024lengthcontrolled,
  author    = {Dubois, Y and Galambosi, B and Liang, P and Hashimoto, T.B},
  title     = {Length-Controlled AlpacaEval: A Simple Way to Debias Automatic Evaluators},
  booktitle = {Conference on Language Modeling (COLM)},
  year      = {2024},
  note      = {{a}rXiv:2404.04475}
}

@inproceedings{zheng2023judging,
  author    = {Zheng, L and Chiang, W.-L and Sheng, Y and Zhuang, S and Wu, Z and Zhuang, Y and Lin, Z and Li, Z and Li, D and Xing, E.P and et~al},
  title     = {Judging LLM-as-a-Judge with MT-Bench and Chatbot Arena},
  booktitle = {Advances in Neural Information Processing Systems (NeurIPS)},
  year      = {2023},
}

@inproceedings{tian2023just,
  author    = {Tian, K and Mitchell, E and Zhou, A and Sharma, A and Rafailov, R and Yao, H and Finn, C and Manning, C.D},
  title     = {Just Ask for Calibration: Strategies for Eliciting Calibrated Confidence Scores from Language Models},
  booktitle = {Conference on Empirical Methods in Natural Language Processing (EMNLP)},
  pages     = {5433--5442},
  year      = {2023},
}

@inproceedings{kuhn2023semantic,
  author    = {Kuhn, L and Gal, Y and Farquhar, S},
  title     = {Semantic Uncertainty: Linguistic Invariances for Uncertainty Estimation in Natural Language Generation},
  booktitle = {International Conference on Learning Representations (ICLR)},
  year      = {2023},
}

@inproceedings{zhai2023sigmoid,
  author    = {Zhai, X and Mustafa, B and Kolesnikov, A and Beyer, L},
  title     = {Sigmoid Loss for Language Image Pre-Training},
  booktitle = {IEEE/CVF International Conference on Computer Vision (ICCV)},
  pages     = {11941--11952},
  year      = {2023},
}

@misc{kaplan2020scaling,
  author    = {Kaplan, J and McCandlish, S and Henighan, T and Brown, T.B and Chess, B and Child, R and Gray, S and Radford, A and Wu, J and Amodei, D},
  title     = {Scaling Laws for Neural Language Models},
  howpublished = {arXiv:2001.08361},
  year      = {2020},
}

@misc{hoffmann2022training,
  author    = {Hoffmann, J and Borgeaud, S and Mensch, A and Buchatskaya, E and Cai, T and Rutherford, E and Casas, D and Hendricks, L.A and Welbl, J and Clark, A and et~al},
  title     = {Training Compute-Optimal Large Language Models},
  booktitle = {Advances in Neural Information Processing Systems (NeurIPS)},
  howpublished = {arXiv:2203.15556},
  year      = {2022},
}

@inproceedings{ruan2024observational,
  author    = {Ruan, Y and Maddison, C.J and Hashimoto, T},
  title     = {Observational Scaling Laws and the Predictability of Language Model Performance},
  booktitle = {Advances in Neural Information Processing Systems (NeurIPS)},
  note      = {{a}rXiv:2405.10938},
  year      = {2024},
}

@misc{wordfreq,
  author    = {Speer, R},
  title     = {wordfreq: Word Frequency Data from Large Corpora},
  howpublished = {\url{https://github.com/rspeer/wordfreq}},
  year      = {2022},
}

@inproceedings{he2016deep,
  author    = {He, K and Zhang, X and Ren, S and Sun, J},
  title     = {Deep Residual Learning for Image Recognition},
  booktitle = {IEEE Conference on Computer Vision and Pattern Recognition (CVPR)},
  pages     = {770--778},
  year      = {2016},
}

@inproceedings{wolfe2022clip,
  author    = {Wolfe, R and Caliskan, A},
  title     = {Markedness in Visual Semantic AI},
  booktitle = {ACM Conference on Fairness, Accountability, and Transparency (FAccT)},
  pages     = {1269--1279},
  year      = {2022},
}

@inproceedings{yao2011stanford40,
  author    = {Yao, B and Jiang, X and Khosla, A and Lin, A.L and Guibas, L and Li, F.-F},
  title     = {Human Action Recognition by Learning Bases of Action Attributes and Parts},
  booktitle = {IEEE International Conference on Computer Vision (ICCV)},
  pages     = {1331--1338},
  year      = {2011},
}

@article{ma2026agreement,
  author    = {Ma, Y and Zhang, L},
  title     = {When Agreement Fails: Visual Anchoring and Multimodal Pseudo-Label Reliability in Classroom Behavior Recognition},
  journal   = {IEEE Access},
  year      = {2026},
  note      = {\url{https://ieeexplore.ieee.org/document/11630593}}
}

@inproceedings{li2023blip2,
  author    = {Li, J and Li, D and Savarese, S and Hoi, S},
  title     = {BLIP-2: Bootstrapping Language-Image Pre-training with Frozen Image Encoders and Large Language Models},
  booktitle = {International Conference on Machine Learning (ICML)},
  pages     = {19730--19742},
  year      = {2023},
}

@article{zhang2023biomedclip,
  author    = {Zhang, S and Xu, Y and Usuyama, N and Bagga, J and Tinn, R and Preston, S and Rao, R and Wei, M and Valluri, N and Wong, C and Lungren, M and Naumann, T and Poon, H},
  title     = {BiomedCLIP: A Multimodal Biomedical Foundation Model Pretrained from Fifteen Million Scientific Image-Text Pairs},
  howpublished = {arXiv:2303.00915},
  year      = {2023},
}

@inproceedings{geifman2017selective,
  author    = {Geifman, Y and El-Yaniv, R},
  title     = {Selective Classification for Deep Neural Networks},
  booktitle = {Advances in Neural Information Processing Systems (NeurIPS)},
  year      = {2017},
}

@article{elyaniv2010foundations,
  author    = {El-Yaniv, R and Wiener, Y},
  title     = {On the Foundations of Noise-Free Selective Classification},
  journal   = {Journal of Machine Learning Research},
  volume    = {11},
  pages     = {1605--1641},
  year      = {2010},
}

@inproceedings{luth2024overcoming,
  author    = {Traub, J and Bungert, T J and L{\"u}th, C T and Baumgartner, M and Maier-Hein, K H and Maier-Hein, L and Jaeger, P F},
  title     = {Overcoming Common Flaws in the Evaluation of Selective Classification Systems},
  booktitle = {Advances in Neural Information Processing Systems (NeurIPS)},
  year      = {2024},
  eprint    = {2407.01032},
  archivePrefix = {arXiv},
}

@inproceedings{srinivasan2024selective,
  author    = {Srinivasan, T and Hessel, J and Gupta, T and Lin, B Y and Choi, Y and Thomason, J and Chandu, K},
  title     = {Selective ``Selective Prediction'': Reducing Unnecessary Abstention in Vision-Language Reasoning},
  booktitle = {Findings of the Association for Computational Linguistics: ACL},
  year      = {2024},
}

\end{document}